\documentclass[10pt,journal,compsoc]{IEEEtran}
\ifCLASSOPTIONcompsoc
  \usepackage[nocompress]{cite}
\else
  \usepackage{cite}
\fi
\ifCLASSINFOpdf
\else
\fi
\usepackage{amsmath}
\usepackage{amsfonts}
\usepackage{graphicx}
\usepackage{algorithm}
\usepackage{algorithmic}
\usepackage{cases}
\usepackage{stfloats}

\newtheorem{thm}{Theorem}
\newtheorem{lema}{Lemma}

\begin{document}
%
\title{Truncated Cauchy Non-negative Matrix Factorization}
%
%
%
%

\author{Naiyang~Guan,~
        Tongliang~Liu,
        Yangmuzi~Zhang,
        Dacheng~Tao,~\IEEEmembership{Fellow,~IEEE}
        and~Larry~S.~Davis,~\IEEEmembership{Fellow,~IEEE}
\IEEEcompsocitemizethanks{
\IEEEcompsocthanksitem N. Guan is with the College of Computer, National University of Defense Technology, Changsha 410073, China. E-mail: ny\_guan@nudt.edu.cn.
\IEEEcompsocthanksitem T. Liu and D. Tao are with the UBTECH Sydney Artificial Intelligence Centre and the School of Information Technologies in the Faculty of Engineering and Information Technologies at University of Sydney, 6 Cleveland St, Darlington, NSW 2008, Australia. E-mail: tongliang.liu@sydney.edu.au, dacheng.tao@sydney.edu.au.
\IEEEcompsocthanksitem Y. Zhang is with Google. E-mail: yangmuzi.zhang@gmail.com.
\IEEEcompsocthanksitem L. Davis is with the University of Maryland Institute for Advanced Computer Studies (UMIACS), University of Maryland at College Park, MD, USA. E-mail: lsd@umiacs.umd.edu. \protect\\
N. Guan and T. Liu contribute equally to this work.
}
}

%
%

\markboth{TPAMI-2017-04-0298.Final, November~2017}%
{Guan \MakeLowercase{\textit{et al.}}: Truncated Cauchy Non-negative Matrix Factorization}
%



\IEEEtitleabstractindextext{%
\begin{abstract}
Non-negative matrix factorization (NMF) minimizes the Euclidean distance between the data matrix and its low rank approximation, and it fails when applied to corrupted data because the loss function is sensitive to outliers. In this paper, we propose a Truncated CauchyNMF loss that handle outliers by truncating large errors, and develop a Truncated CauchyNMF to robustly learn the subspace on noisy datasets contaminated by outliers. We theoretically analyze the robustness of Truncated CauchyNMF comparing with the competing models and theoretically prove that Truncated CauchyNMF has a generalization bound which converges at a rate of order $O(\sqrt{{\ln n}/{n}})$, where $n$ is the sample size. We evaluate Truncated CauchyNMF by image clustering on both simulated and real datasets. The experimental results on the datasets containing gross corruptions validate the effectiveness and robustness of Truncated CauchyNMF for learning robust subspaces.
\end{abstract}

}

\maketitle

\IEEEdisplaynontitleabstractindextext

%
\IEEEpeerreviewmaketitle

\IEEEraisesectionheading{\section{Introduction}\label{sec:introduction}}

%
%
%
%
\IEEEPARstart{N}{on-negative} matrix factorization (NMF, \cite{bib26}) explores the non-negativity property of data and has received considerable attention in many fields, such as text mining \cite{bib38}, hyper-spectral imaging \cite{bib39}, and gene expression clustering \cite{bib60}. It decomposes a data matrix into the product of two lower dimensional non-negative factor matrices by minimizing the Eudlidean distance between their product and the original data matrix. Since NMF only allows additive, non-subtractive combinations, it obtains a natural parts-based representation of the data. NMF is optimal when the dataset contains additive Gaussian noise, and so it fails on grossly corrupted datasets, e.g., the AR database \cite{bib33} where face images are partially occluded by sunglasses or scarves. This is because the corruptions or outliers seriously violate the noise assumption.

Many models have been proposed to improve the robustness of NMF. Hamza and Brady \cite{bib18} proposed a hypersurface cost based NMF (HCNMF) which minimizes the hypersurface cost function\footnote{The hypersurface cost function is defined as $h(x)=\sqrt{(1+x^2)}-1$ which is quadratic when its argument is small and linear when its argument is large.} between the data matrix and its approximation. HCNMF is a significant contribution for improving the robustness of NMF, but its optimization algorithm is time-consuming because the Armijo's rule based line search that it employs is complex. Lam \cite{bib25} proposed $L_1$-NMF\footnote{When the noise is modeled by Laplace distribution, the maximum likelihood estimation yields an $L_1$-norm based objective function. We therefore term the method in \cite{bib25} $L_1$-NMF.} to model the noise in a data matrix by a Laplace distribution. Although $L_1$-NMF is less sensitive to outliers than NMF, its optimization is expensive because the $L_1$-norm based loss function is non-smooth. This problem is largely reduced by Manhattan NMF (MahNMF, \cite{bib17}), which solves $L_1$-NMF by approximating the non-smooth loss function with a smooth one and minimizing the approximated loss function with Nesterov's method \cite{bib58}. Zhang \textit{et al}. \cite{bib48} proposed an $L_1$-norm regularized Robust NMF (RNMF-$L_1$) to recover the uncorrupted data matrix by subtracting a sparse error matrix from the corrupted data matrix. Kong \textit{et al}. \cite{bib24} proposed $L_{2,1}$-NMF to minimize the $L_{2,1}$-norm of an error matrix to prevent noise of large magnitude from dominating the objective function. Gao \textit{et al}. \cite{bib70} further proposed robust capped norm NMF (RCNMF) to filter out the effect of outlier samples by limiting their proportions in the objective function. However, the iterative algorithms utilized in $L_{2,1}$-NMF and RCNMF converge slowly because they involve a successive use of the power method \cite{bib2}. Recently, Bhattacharyya \textit{et al}. \cite{bib71} proposed an important robust variant of convex NMF which only requires the average $L_1$-norm of noise over large subsets of columns to be small; Pan \textit{et al}. \cite{bib72} proposed an $L_1$-norm based robust dictionary learning model; and Gillis and Luce \cite{bib73} proposed a robust near-separable NMF which can determine the low-rank, avoid normalizing data, and filter out outliers. HCNMF, $L_1$-NMF, RNMF-$L_1$, $L_{2,1}$-NMF, RCNMF, \cite{bib71}, \cite{bib72} and \cite{bib73} share a common drawback, i.e., they all fail when the dataset is contaminated by serious corruptions because the breakdown point of the $L_1$-norm based models is determined by the dimensionality of the data \cite{bib9}.

In this paper, we propose a Truncated Cauchy non-negative matrix factorization (Truncated CauchyNMF) model to learn a subspace on a dataset contaminated by large magnitude noise or corruption. In particular, we proposed a Truncated Cauchy loss that simultaneously and appropriately models moderate outliers (because the loss corresponds to a fat tailed distribution in-between the truncation points) and extreme outliers (because the truncation directly cut off large errors). Based on the proposed loss function, we develop a novel Truncated CauchyNMF model. We theoretically analyze the robustness of Truncated CauchyNMF and show that Truncated CauchyNMF is more robust than a family of NMF models, and derive a theoretical guarantee for its generalization ability and show that Truncated CauchyNMF converges at a rate of order $O(\sqrt{{\ln n}/{n}})$, where $n$ is the sample size. Truncated CauchyNMF is difficult to optimize because the loss function includes a nonlinear logarithmic function. To address this, we optimize Truncated CauchyNMF by half-quadratic (HQ) programming based on the theory of convex conjugation. HQ introduces a weight for each entry of the data matrix and alternately and analytically updates the weight and updates both factor matrices by easily solving a weighted non-negative least squares problem with Nesterov's method \cite{bib35}. Intuitively, the introduced weight reflects the magnitude of the error. The heavier the corruption, the smaller the weight, and the less an entry contributes to learning the subspace. By performing truncation on magnitudes of errors, we prove that HQ introduces zero weights for entries with extreme outliers, and thus HQ is able to learn the intrinsic subspace on the inlier entries.

In summary, the contributions of this paper are three-fold: (1) we propose a robust subspace learning framework called Truncated CauchyNMF, and develop a Nesterov-based HQ algorithm to solve it; (2) we theoretically analyze the robustness of Truncated CauchyNMF comparing with a family of NMF models, and provide insight as to why Truncated CauchyNMF is the most robust method; and (3) we theoretically analyze the generalization ability of Truncated CauchyNMF, and provide performance guarantees for the proposed model. We evaluate Truncated CauchyNMF by image clustering on both simulated and real datasets. The experimental results on the datasets containing gross corruptions validate the effectiveness and robustness of Truncated CauchyNMF for learning the subspace.

The rest of this paper is organized as follows: Section \ref{sec:chapter2} describes the proposed Truncated CauchyNMF, Section \ref{sec:chapter3} develops the Nesterov-based half-quadratic (HQ) programming algorithm for solving Truncated CauchyNMF. Section \ref{sec:chapter4} surveys the related works and Section \ref{sec:chapter5} verifies Truncated CauchyNMF on simulated and real datasets. Section \ref{sec:chapter6} concludes this paper. All the proofs are given in the supplementary material.

%
%

\hfill

\hfill

\section{Truncated Cauchy Non-negative Matrix Factorization}\label{sec:chapter2}
Classical NMF \cite{bib26} is not robust because its loss function $e_2(x)=x^2$ is sensitive to outliers considering the errors of large magnitude dominate the loss function. Although some robust loss functions, such as $e_1(x)=|x|$ for $L_1$-NMF \cite{bib25}, Hypersurface cost $e_h(x)=\sqrt{1+x^2}-1$ \cite{bib18}, and Cauchy loss $e_c(x;\gamma)=\ln\left(1+\left({x}/{\gamma}\right)^2\right)$, are less sensitive to outliers, they introduces infinite energy for infinitely large noise in the extreme case. To remedy this problem, we propose a Truncated Cauchy loss by truncating the magnitudes of large errors to limit the effects of extreme outliers, i.e.,
\begin{equation}
\label{eqn2-1}
e_t(x;\gamma,\varepsilon)=\left\{\begin{array}{cc}\ln\left(1+\left({x}/{\gamma}\right)^2\right),&|x|\le\varepsilon\\\ln\left(1+\left({\varepsilon}/{\gamma}\right)^2\right),&|x|>\varepsilon\end{array}\right.,
\end{equation}
where $\gamma$ is the scale parameter of the Cauchy distribution and $\varepsilon$ is a constant.

\begin{figure}[!t]
\centering
\includegraphics[width=1.0\linewidth]{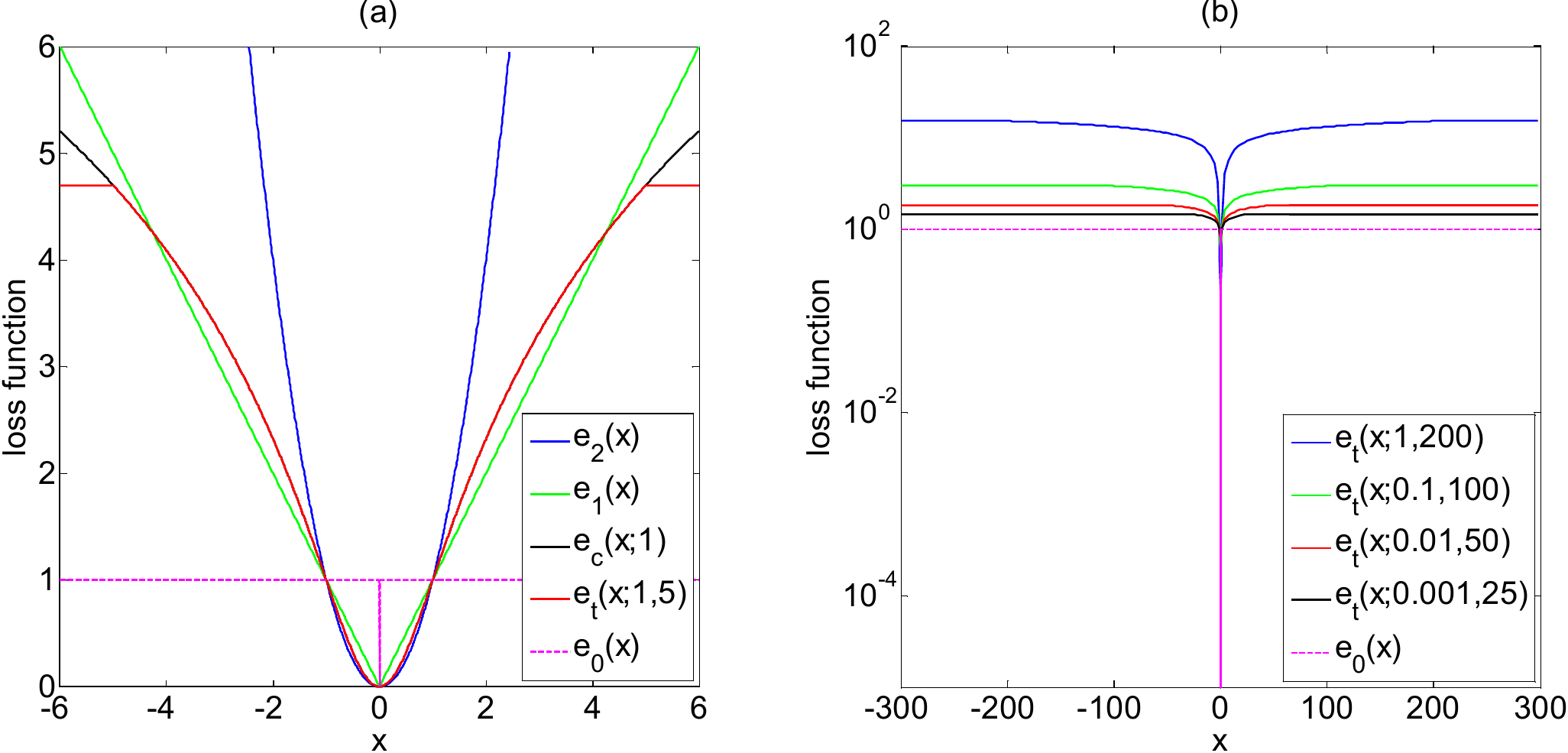}
\caption{The comparison of loss functions: (a) $e_2(x)$, $e_1(x)$, $e_c(x;1)$, $e_t(x;1,5)$, and $e_0(x)$; and (b) $e_t(x;\gamma,\varepsilon$) when $(\gamma,\varepsilon)=(1,200),(0.1,100),(0.01,50),(0.001,25)$ and $e_0(x)$.}
\label{fig1-1}
\end{figure}

To study the behavior of the Truncated Cauchy loss, we compare the loss functions $e_2(x)$, $e_1(x)$, $e_c(x;1)$, $e_t(x;1,5)$, and the loss function of the $L_0$-norm, i.e., $e_0(x)=\left\{\begin{array}{cc}1,&x\neq0\\0,&x=0\end{array}\right.$ in Figure \ref{fig1-1}, because the $L_0$-norm induces robust models. Figure \ref{fig1-1}(a) shows that when the error is moderately large, e.g., $|x|\le 5$, $e_t(x;1,5)$ shifts from $e_2(x)$ to $e_1(x)$ and corresponds to a fat-tailed distribution, and implies that the Truncated Cauchy loss can model moderate outliers well, while $e_2(x)$ cannot because it makes the outliers dominate the objective function. When the error gets larger and larger, $e_t(x;1,5)$ gets away from $e_1(x)$ and behaves like $e_0(x)$, and $e_t(x;1,5)$ keeps constant once the error exceeds a threshold, e.g., $|x|>5$, and implies that the Truncated Cauchy loss can model extreme outliers, whereas neither $e_1(x)$ nor $e_c(x;1)$ cannot because they encourage infinite energy to infinitely large error. Intuitively, the Truncated Cauchy loss can model both moderate and extreme outliers well. Figure \ref{fig1-1}(b) plots the curves of both $e_t(x;\gamma,\varepsilon)$ and $e_0(x)$ with varying $\gamma$ from $0.001$ to $1$ and accordingly varying $\varepsilon$ from $25$ to $200$. It shows that $e_t(x;\gamma,\varepsilon)$ behaves more and more close to $e_0(x)$ when $\gamma$ approaches zero. By comparing the behaviors of loss functions, we believe that the Truncated Cauchy loss can induce robust NMF model.

Given $n$ high-dimensional samples arranged in a non-negative matrix $V=[v_1,\dots,v_n]\in \mathbb{R}_+^{m\times n}$, Truncated Cauchy non-negative matrix factorization (Truncated CauchyNMF) approximately decomposes $V$ into the product of two lower dimensional non-negative matrices, i.e., $V=WH+E$, where $W\in\mathbb{R}_+^{m\times r}$ signifies the basis, $H=[h_1,\dots,h_n]\in\mathbb{R}_+^{r\times n}$ signifies the coefficients, and $E\in\mathbb{R}^{m\times n}$ signifies the error matrix which is measured by using the proposed Truncated Cauchy loss. The objective function of Truncated CauchyNMF can be written as
\begin{equation}
\label{eqn2-4}
\min_{W\ge0,H\ge0}\frac{1}{2}\sum_{ij}\mathit{g}((\frac{V-WH}{\gamma})_{ij}^2),
\end{equation}
where $\mathit{g}(x)=\left\{\begin{array}{cc}\ln(1+x),&0\le x\le \sigma\\\ln(1+\sigma),&x>\sigma\end{array}\right.$ is utilized for the convenience of derivation and $\sigma$ is a truncation parameter, and $\gamma$ is the scale parameter. We will next show that the truncation parameter $\sigma$ can be implicitly determined by robust statistics and the scale parameter $\gamma$ can be estimated by the Nagy algorithm \cite{bib51}. It is not hard to see that Truncated CauchyNMF includes CauchyNMF as a special case when $\sigma=+\infty$. Since \eqref{eqn2-4} assigns fixed energy to any large error whose magnitude exceeds $\gamma\sqrt{\sigma}$, Truncated CauchyNMF can filter out any extreme outliers.

\begin{figure}[!b]
\centering
\includegraphics[width=1.0\linewidth]{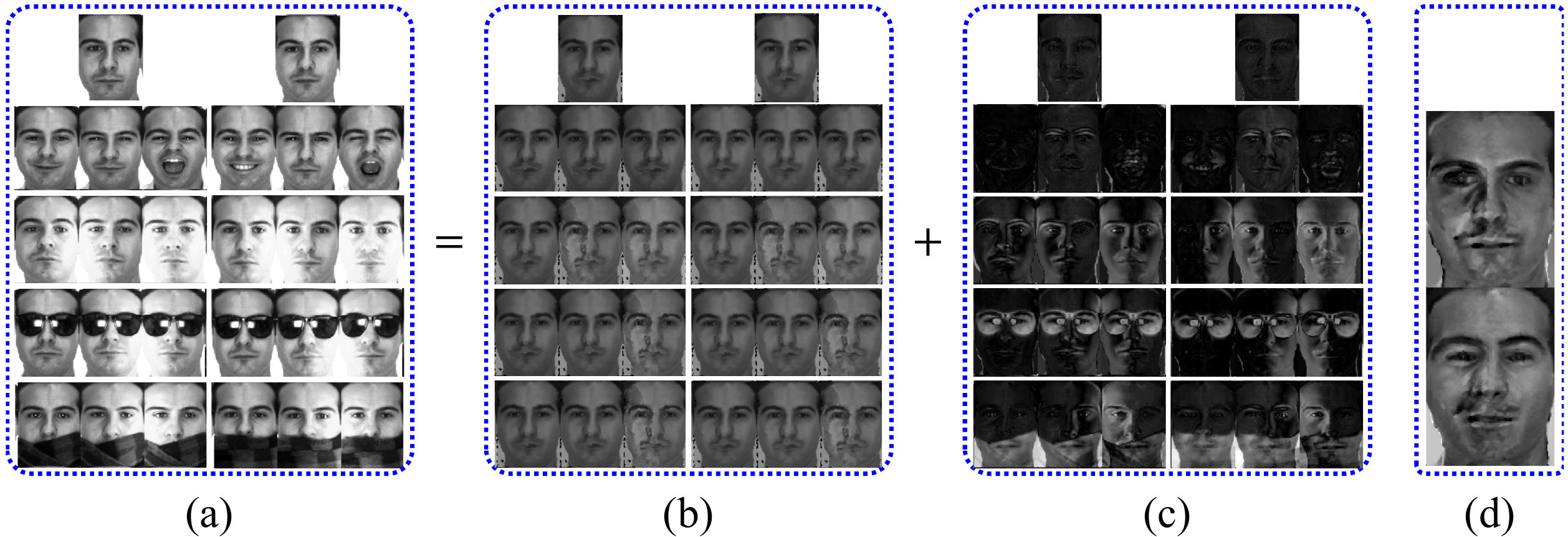}
\caption{The illustrative example: (a) frontal face images from the AR database, (b) face images reconstructed by Truncated CauchyNMF, (c) error images, and (d) the learned basis images.}
\label{fig1-2}
\end{figure}

To illustrate the ability of Truncated CauchyNMF to model outliers, Figure \ref{fig1-2} gives an illustrative example that demonstrates its application to corrupted face images. In this example, we select $26$ frontal face images of an individual in two sessions from the Purdue AR database \cite{bib33} (see all face images in Figure \ref{fig1-2}(a)). In each session, there are $13$ frontal face images with different facial expressions, captured under different illumination conditions, with sunglasses, and with a scarf. Each image is cropped into a $165\times120$-dimensional pixel array and reshaped into a $19800$-dimensional vector. The total number of face images compose a $19800\times26$-dimensional non-negative matrix because the pixel values are non-negative. In this experiment, we aim at learning the intrinsically clean face images from the contaminated images. This task is quite challenging because more than half the images are contaminated. Since these images were taken in two sessions, we set the dimensionality low ($r=2$) to learn two basis images. Figure \ref{fig1-2}(b) shows that Truncated CauchyNMF robustly recovers all face images even when they are contaminated by a variety of facial expressions, illumination, and occlusion. Figure \ref{fig1-2}(c) presents the reconstruction errors and Figure \ref{fig1-2}(d) shows the basis images, which confirms that Truncated CauchyNMF is able to learn clean basis images with the outliers filtered out.

\begin{figure*}[hb]
\hrulefill
\begin{eqnarray}
\label{eqn-thm1}
&&R(f_{W_n})-R(f_{W_*})\le\sup_{f_W\in F_\mathcal{W}}\left|E_v\frac{1}{2n}\sum_{ij}\mathit{g}\left(\left(\frac{V-WH}{\gamma}\right)_{ij}^2\right)-\frac{1}{2n}\sum_{ij}\mathit{g}\left(\left(\frac{V-WH}{\gamma}\right)_{ij}^2\right)\right|\nonumber\\
&&\le \min_{\epsilon}\left\{2\epsilon+\frac{\alpha^2}{\gamma^2}\sqrt{\left({mr\ln\left(({4^{\frac{1}{m}}\pi^{\frac{1}{2}}\left(\frac{8r\alpha^2+2\alpha^2}{r^2}+\frac{\sigma\alpha^2}{\sqrt{2}r^3}+\frac{2r\sigma^2\alpha^2}{r^4}\right)mr})/{\Gamma(\frac{m}{2})^{\frac{1}{m}}2\epsilon}\right)+\ln(\frac{2}{\delta})}\right)/{2n}}\right\}\nonumber\\
&&\le \frac{2}{n}+\frac{\alpha^2}{\gamma^2}\sqrt{\left({mr\ln\left(\left({4^{\frac{1}{m}}\pi^{\frac{1}{2}}\left(\frac{8r\alpha^2+2\alpha^2}{r^2}+\frac{\sigma\alpha^2}{\sqrt{2}r^3}+\frac{2r\sigma^2\alpha^2}{r^4}\right)mrn}\right)/{\Gamma(\frac{m}{2})^{\frac{1}{m}}2}\right)+\ln(\frac{2}{\delta})}\right)/{2n}}.
\end{eqnarray}
\end{figure*}

In the following subsections, we will analyze the generalization ability and robustness of Truncated CauchyNMF. Before that, we introduce \textbf{\textit{Lemma 1}} which states that the new representations generated by Truncated CauchyNMF are bounded if the input observations are bounded. This lemma will be utilized in the following analysis with the only assumption that each base is a unit vector. Such an assumption is typical in NMF because the bases $W$ are usually normalized to limit the variance of its local minimizers. We use $\|\cdot\|_p$ to represent the $L_p$-norm and $\|\cdot\|$ to represent the Euclidean norm.

\begin{lema}
Assuming $\|W_i\|=1$, $i=1,\dots,r$, and that the input observations are bounded, i.e., $\|v\|\le \alpha$ for some $\alpha>0$. Then the new representations are also bounded, i.e., $\|h\|\le 2\alpha+(\sigma \alpha)/(\sqrt{2}\gamma)$.
\end{lema}

Although Truncated CauchyNMF \eqref{eqn2-4} has a differentiable objective function, solving it is difficult because the natural logarithmical function is nonlinear. Section \ref{sec:chapter3} will present a half-quadratic (HQ) programming algorithm for solving Truncated CauchyNMF.

\subsection{Generalization Ability}
To analyze the generalization ability of Truncated CauchyNMF, we further assume that samples $[v_1,\dots,v_n]$ are independent and identically distributed and drawn from a space $\mathcal{V}$ with a Borel measure $\rho$. We use $A_{\cdot j}$ and $A_{ij}$ to denote the $j$-th column and the $(i,j)$-th entry of a matrix , respectively, and $a_i$ is the $i$-th entry of a vector $a$.

For any $W\in\mathbb{R}_+^{m\times r}$, we define the reconstruction error of a sample $v$ as follows:
\begin{equation}
\label{eqn2-5}
f_W(v)=\min_{h\in\mathbb{R}_+^r}\sum_{j}\mathit{g}((\frac{v-Wh}{\gamma})_j^2).
\end{equation}
Therefore, the objective function of Truancated CauchyNMF \eqref{eqn2-4} can be written as
\begin{equation}
\label{eqn2-6}
\min_{W\ge0,H\ge0}\frac{1}{2}\sum_{ij}\mathit{g}((\frac{V-WH}{\gamma})_{ij}^2)=\min_{W\ge0}\frac{1}{2}\sum_if_W(v_i).
\end{equation}
Let us define the empirical reconstruction error of Truncated CauchyNMF as $R_n(f_W)=\frac{1}{n}\sum_{i=1}^nf_W(v_i)$, and the expected reconstruction error of Truncated CauchyNMF as $R(f_W)=E_v\frac{1}{n}\sum_{i=1}^nf_W(v_i)$. Intuitively, we want to learn
\begin{equation}
\label{eqn2-7}
W_*={\arg\min}_{W\geq 0}R(f_W).
\end{equation}
However, since the distribution of $v$ is unknown, we cannot minimize $R(f_W)$ directly. Instead, we use the empirical risk minimization (ERM, \cite{bib3}) algorithm to learn $W_n$ to approximate $W_*$, as follows:
\begin{equation}
\label{eqn2-8}
W_n={\arg\min}_{W\geq 0}R_n(f_W).
\end{equation}

We are interested in the difference between $W_n$ and $W_*$. If the distance is small, we can say that $W_n$ is a good approximation of $W_*$. Here, we measure the distance of their reduced expected reconstruction error as follows:
\begin{eqnarray*}
R(f_{W_n})-R(f_{W_*})
\le2\sup_{f_W\in F_\mathcal{W}}|R(f_W)-R_n(f_W)|,
\end{eqnarray*}
where $F_\mathcal{W}=\{f_W|W\in\mathcal{W}=\mathbb{R}_+^{m\times r}\}$. The right hand side is known as the generalization error. Note that since NMF is convex with respect to either $W$ or $H$ but not both, the minimizer $f_{W_n}$ is hard to obtain. In practice, a local minimizer is used as an approximation. Measuring the distance between the local minimizer and the global minimizer is also an interesting and challenging problem.

By analyzing the covering number \cite{bib49} of the function class $F_\mathcal{W}$ and \textbf{\textit{Lemma 1}}, we derive a generalization error bound for Truncated CauchyNMF as follows:
\begin{thm}
\label{thm1-1}
Let $\|W_{\cdot i}\|=1$, $i=1,\dots,r$, and $F_\mathcal{W}=\{f_W|W\in \mathcal{W}=\mathbb{R}_+^{m\times r}\}$. Assume that $\|v\|\le \alpha$. For any $\delta>0$, with probability at least $1-\delta$, the equation \eqref{eqn-thm1} holds, where $\Gamma(\frac{1}{2})=\sqrt{\pi}$; $\Gamma(1)=1$; and $\Gamma(x+1)=x\Gamma(x)$.
\end{thm}

\textit{\textbf{Remark 1}}. \textbf{Theorem \ref{thm1-1}} shows that under the setting of our proposed Truncated CauchyNMF, the expected reconstruction error $R(f_{W_n})$ will converge to $R(f_{W_*})$ with a fast rate of order $O(\sqrt{{\ln n}/{n}})$, which means that when the sample size $n$ is large, the distance between $R(f_{W_n})$ and $R(f_{W_*})$ will be small. Moreover, if $n$ is large and a local minimizer $W$ (obtained by optimizing the non-convex objective of Truncated CauchyNMF) is close to the global minimizer $W_n$, the local minimizer will also be close to the optimal $W_*$.

\textit{\textbf{Remark 2}}. \textbf{Theorem \ref{thm1-1}} also implies that for any $W$ learned from (\ref{eqn2-4}), the corresponding empirical reconstruction error $R_n(f_W)$ will converge to its expectation with a specific rate guarantee, which means our proposed Truncated CauchyNMF can generalize well to unseen data.

Note that the noise sampled from the Cauchy distribution should not be bounded because Cauchy distribution is heavy-tailed. And bounded observations always imply bounded noise. However, \textbf{Theorem \ref{thm1-1}} keeps the boundedness assumption on the observations for two reasons: (1) the truncated loss function indicates that the observations corresponding to unbounded noise are discarded, and (2) in real applications, the energy of observations should be bounded, which means their $L_2$-norms are bounded.

\subsection{Robustness Analysis}
We next compare the robustness of Truncated CauchyNMF with those of other NMF models by using a sample-weighted procedure interpretation \cite{bib31}. The sample-weighted procedure compares the robustness of different algorithms from the optimization viewpoint.

Let $F(WH)$ denote the objective function of any NMF problem and $f(t)=F(tWH)$ where $t\in\mathbb{R}$. We can verify that the NMF problem is equivalent to finding a pair of $WH$ such that $f^{'}(1)=0$\footnote{When minimizing $F(WH)$, the low rank matrices $W$ and $H$ will be updated alternately. Fixing one of them and optimizing the other implies that $f^{'}(1)=0$. In other words, if $f^{'}(1)\neq 0$, neither $W$ nor $H$ can be a minimizer.}, where $f^{'}(t)$ denotes the derivative of $f(t)$. Let $c(V_{ij},WH)=(V-WH)_{ij}(-WH)_{ij}$ be the contribution of the $j$-th entry of the $i$-th training example to the optimization procedure and $e(V_{ij},WH)=|V-WH|_{ij}$ be an error function. Note that we choose $c(V_{ij},WH)$ as the basis of contribution because we choose NMF, which aims to find a pair of $WH$ such that $\sum_{ij}c(V_{ij},WH)=0$ and is sensitive to noise, as the baseline for comparing the robustness. Also note that $e(V_{ij},WH)$ represents the noise added to the $(i,j)$-th entry of $V$. The interpretation of the sample-weighted procedure explains the optimization procedure as being contribution-weighted with respect to the noise.

\begin{table*}[!t]
\caption{Comparison of the robustness of Truncated CauchyNMF with those of other NMF models.}
\label{tbl2-1}
\centering
\begin{tabular}{c||ccc}
\hline
NMF methods &Objective function $F(WH)$  &Derivative $f'(1)$\\
\hline
NMF  &$\|V-WH\|_F^2$ &$\sum_{ij}2c(V_{ij},WH)$\\
HCNMF  &$\sum_{ij}(\sqrt{1+(V-WH)_{ij}^2}-1)$ &$\sum_{ij}\frac{1}{\sqrt{1+(V-WH)_{ij}^2}}c(V_{ij},WH)$\\
$L_{2,1}$-NMF  &$\|V-WH\|_{2,1}$ &$\sum_{ij}\frac{1}{\sqrt{\sum_l(V-WH)_{lj}^2}}c(V_{ij},WH)$\\
RCNMF  &$\sum_{j=1}^n\min\{\|V_{\cdot j}-WH_{\cdot j}\|,\theta\}$ &$\sum_{j=1}^n\left\{\begin{array}{cc}\sum_i\frac{1}{\sqrt{\sum_l(V-WH)_{lj}^2}}c(V_{ij},WH),&\|V_{\cdot j}-WH_{\cdot j}\|\le\theta\\0,&\|V_{\cdot j}-WH_{\cdot j}\|\ge\theta\end{array}\right.$\\
RNMF-$L_1$  &$\|V-WH-S\|_F^2+\lambda\|S\|_1$ &$\sum_{ij}2(1-\frac{S_{ij}}{(V-WH)_{ij}})c(V_{ij},WH)$\\
$L_1$-NMF  &$\|V-WH\|_1$ &$\sum_{ij}\frac{1}{|V-WH|_{ij}}c(V_{ij},WH)$\\
HuberNMF  &$\begin{array}{c}\sum_{i=1}^m\sum_{j=1}^n\mathit{l}((V-WH)_{ij},\sigma),\\where\;\mathit{l}(x,\sigma)=\left\{\begin{array}{cc}x^2,&|x|\le\sigma\\2\sigma|x|-\sigma^2,&|x|\ge\sigma\end{array}\right.\end{array}$ &$\sum_{ij}\left\{\begin{array}{cc}2c(V_{ij},WH),&|V-WH|_{ij}\le\sigma\\\frac{2\sigma}{|V-WH|_{ij}}c(V_{ij},WH),&|V-WH|_{ij}\ge\sigma\end{array}\right.$\\
CIM-NMF &$\sum_{i=1}^m\sum_{j=1}^n1-\frac{1}{\sqrt{2\pi}\sigma}e^{-{(V-WH)_{ij}^2}/{2\sigma^2}}$  &$\sum_{ij}\frac{1}{\sqrt{2\pi}\sigma^3e^{-{(V-WH)_{ij}^2}/{2\sigma^2}}}c(V_{ij},WH)$\\
CauchyNMF &$\sum_{ij}\ln(1+(\frac{V-WH}{\gamma})_{ij})$  &$\sum_{ij}\frac{2}{\gamma^2+(V-WH)_{ij}^2}c(V_{ij},WH)$\\
Truncated CauchyNMF &$\begin{array}{c}\sum_{ij}\mathit{g}((\frac{V-WH}{\gamma})_{ij}^2),\\where\;\mathit{g}(x)=\left\{\begin{array}{cc}\ln(1+x),&0\le x\le\sigma\\\ln(1+\sigma),&x>\sigma\end{array}\right.\end{array}$ &$\sum_{ij}\left\{\begin{array}{cc}\frac{2}{\gamma^2+(V-WH)_{ij}^2}c(V_{ij},WH),&|V-WH|_{ij}\le\gamma\sqrt{\sigma}\\0\cdot c(V_{ij},WH),&|V-WH|_{ij}>\gamma\sqrt{\sigma}\end{array}\right.$\\
\hline
\end{tabular}
\end{table*}

We compare $f'(1)$ of a family of NMF models in Table \ref{tbl2-1}. Note that since multiplying $f'(1)$ by a constant will not change its zero points, we can normalize the weights of different NMF models to unity when the noise is equal to zero. During the optimization procedures, robust algorithms should assign a small weight to an entry of the training set with large noise. Therefore, by comparing the derivative $f'(1)$, we can easily make the following statements: (1) $L_1$-NMF\footnote{For the soundness of defining the subgradient of $L_1$-norm, we state that $\frac{0}{0}$ can be any value in $[-1,1]$.} is more robust to noise and outliers than NMF; Huber-NMF combines the ideas of NMF and $L_1$-NMF; (2) HCNMF, $L_{2,1}$-NMF, RCNMF, and RNMF-$L_1$ work similarly to $L_1$-NMF because their weights are of order $O({1}/{e(V_{ij},WH)})$ with respect to the noise. It also becomes clear that HCNMF, $L_{2,1}$-NMF, and RCNMF exploit some data structure information because the weights include the neighborhood information of $e(V_{ij},WH)$ and that RNMF-$L_1$ is less sensitive to noise because it employs a sparse matrix $S$ to adjust the weights; (3) The interpretation of the sample-weighted procedure also illustrates why CIM-NMF works well for heavy noise. This is because its weights decrease exponentially when the noise is large; And (4) for the proposed Truncated CauchyNMF, when the noise is larger than a threshold, its weights will drop directly to zero, which decrease far faster than that of CIM-NMF and thus Truncated CauchyNMF is very robust to extreme outliers. Finally, we conclude that Truncated CauchyNMF is more robust than any other NMF models with respect to extreme outliers because it has the power to provide smaller weights to examples.

\section{Half-Quadratic Programming Algorithm for Truncated CauchyNMF}
\label{sec:chapter3}
Note that Truncated CauchyNMF \eqref{eqn2-4} cannot be solved directly because the energy function $\mathit{g}(x)$ is non-quadratic. We present a half-quadratic (HQ) programming algorithm based on conjugate function theory [9]. To adopt the HQ algorithm, we transform \eqref{eqn2-4} to the following maximization form:
\begin{equation}
\label{eqn3-1}
\max_{W\ge0,H\ge0}\frac{1}{2}\sum_{ij}f((\frac{V-WH}{\gamma})_{ij}^2),
\end{equation}
where $f(x)=-\mathit{g}(x)$ is the core function utilized in HQ. Since the negative logarithmic function is convex, $f(x)$ is also convex.

\subsection{HQ-based Alternating Optimization}
Generally speaking, the half-quadratic (HQ) programming algorithm \cite{bib13} reformulates the non-quadratic loss function as an augmented loss function in an enlarged parameter space by introducing an additional auxiliary variable based on the convex conjugation theory \cite{bib4}. HQ is equivalent to the quasi-Newton method \cite{bib36} and has been widely applied in non-quadratic optimization.

Note that the function $f(x):\mathbb{R}_+\rightarrow\mathbb{R}$ is continuous, and according to \cite{bib4}, its conjugate $f^*(y):\mathbb{R}\rightarrow\mathbb{R}\cup\{+\infty\}$ is defined as
\begin{equation*}
f^*(y)=\max_{x\in\mathbb{R}_+}\{xy-f(x)\}.
\end{equation*}
Since $f(x)$ is convex and closed (although the domain $\mathbb{R}_+$ is open, $f(x)$ is closed, see Section A.3.3 in \cite{bib4}), the conjugate of its conjugate function is itself \cite{bib4}, i.e., $f^{**}=f$, then we have:
\begin{thm}
\label{thm4-1}
The core function $f(x)$ and its conjugate $f^*(y)$ satisfy
\begin{equation}
\label{eqn3-2}
f(x)=\max_y\{yx-f^*(y)\},x\in\mathbb{R}_+,
\end{equation}
and the maximizer is $y_*=\left\{\begin{array}{cc}-{1}/{(1+x)},&0\le x\le \sigma\\0,&x>\sigma\end{array}\right.$.
\end{thm}

By substituting $x=(\frac{V-WH}{\gamma})_{ij}^2$ into \eqref{eqn3-2}, we have the augmented loss function
\begin{equation}
\label{eqn3-3}
f((\frac{V-WH}{\gamma})_{ij}^2)=\max_{Y_{ij}}\{Y_{ij}(\frac{V-WH}{\gamma})_{ij}^2-f^*(Y_{ij})\},
\end{equation}
where $Y_{ij}$ is the auxiliary variable introduced by HQ for $(\frac{V-WH}{\gamma})_{ij}^2$. By substituting \eqref{eqn3-3} into \eqref{eqn3-1}, we have the objective function in an enlarged parameter space
\begin{eqnarray}
\label{eqn3-4}
&&\max_{W\ge0,H\ge0}\{\frac{1}{2}\sum_{ij}\max_{Y_{ij}}\{Y_{ij}(\frac{V-WH}{\gamma})_{ij}^2-f^*(Y_{ij})\}\}=\nonumber\\
&&\max_{W\ge0,H\ge0,Y}\{\frac{1}{2}\sum_{ij}\{Y_{ij}(\frac{V-WH}{\gamma})_{ij}^2-f^*(Y_{ij})\}\},
\end{eqnarray}
where the equality comes from the separability of the optimization problems with respect to $Y_{ij}$.

Although the objective function in \eqref{eqn3-1} is non-quadratic, its equivalent problem \eqref{eqn3-4} is essentially a quadratic optimization. In this paper, HQ solves (\ref{eqn3-4}) based on the block coordinate descent framework. In particular, HQ recursively optimizes the following three problems. At $t$-th iteration,
\begin{align}
Y^{t+1}:\;&{\max}_Y \frac{1}{2} \sum_{ij}(Y_{ij}(\frac{V-W^tH^t}{\gamma})_{ij}^2-f^*(Y_{ij})),\label{eqn3-5}\\
H^{t+1}:\;&{\max}_{H\ge0} \frac{1}{2} \sum_{ij}(Y_{ij}^{t+1} (\frac{V-W^tH}{\gamma})_{ij}^2),\label{eqn3-6}\\
W^{t+1}:\;&{\max}_{W\ge0} \frac{1}{2} \sum_{ij}(Y_{ij}^{t+1} (\frac{V-WH^{t+1}}{\gamma})_{ij}^2).\label{eqn3-7}
\end{align}
Using \textbf{Theorem \ref{thm4-1}}, we know that the solution of \eqref{eqn3-5} can be expressed analytically as
\begin{equation*}
Y_{ij}^{t+1}=\left\{\begin{array}{cc}-\frac{1}{1+(\frac{V-W^tH^t}{\gamma})_{ij}^2},&if\;|(V-W^tH^t)_{ij}|\le\gamma\sqrt{\sigma}\\0,&if\;|(V-W^tH^t)_{ij}|>\gamma\sqrt{\sigma}\end{array}\right..
\end{equation*}

Since \eqref{eqn3-6} and \eqref{eqn3-7} are symmetric and intrinsically weighted non-negative least squares (WNLS) problems, they can be optimized in the same way using the Nesterov method \cite{bib16}. Taking \eqref{eqn3-6} as an example, the procedure of its Nesterov based optimization is summarized in \textbf{Algorithm \ref{alg3-1}}, and its derivative is derived in the supplementary material. Considering that \eqref{eqn3-6} is a constrained optimization problem, similar to \cite{bib28}, we use the following projected gradient-based criterion to check the stationarity of the search point , i.e., $\nabla_j^P(h_k)=0$, where $\nabla_j^P(h_k)_l=\left\{\begin{array}{cc}\nabla_j^P(h_k)_l,&(h_k)_l\ge 0\\\min\{0,\nabla_j^P(h_k)_l\},&(h_k)_l=0\end{array}\right.$. Since the above stopping criterion will make OGM run unnecessarily long, similar to \cite{bib28}, we use a relaxed version
\begin{equation}
\label{eqn3-8}
\|\nabla_j^P(h_k)\|_F\le \max\{\epsilon_1,10^{-3}\}\times\|\nabla_j^P(h_0)\|_F,
\end{equation}
where $\epsilon_1$ is a tolerance that controls how far the search point is from a stationary point.

\begin{algorithm}
\caption{Optimal Gradient Method (OGM) for WNLS}
\label{alg3-1}
\begin{algorithmic}
\STATE \textbf{Input}: $V_{\cdot j}\in\mathbb{R}_+^m$, $W^t\in\mathbb{R}_+^{m\times r}$, $H_{\cdot j}^t\in\mathbb{R}_+^r$, $D_j^{t+1}$.
\STATE \textbf{Output}: $H_{\cdot j}^{t+1}$.
\STATE 1: Initialize $z^0=H_{\cdot j}^t$, $h^0=H_{\cdot j}^t$, $\alpha_0=1$, $k=0$.
\STATE 2: Calculate $L_j=\|{W^t}^TD_j^{t+1}W^t\|_2$.
\REPEAT
\STATE 3: $\nabla_j (z^k)={W^t}^T D_j^{t+1} W^t z^k-{W^t}^T D_j^{t+1} V_{\cdot j}$.
\STATE 4: $h^{k+1}=\Pi_+(z^k-\frac{\nabla_j(z^k)}{L_j})$.
\STATE 5: $\alpha_{k+1}=\frac{1+\sqrt{4\alpha_k^2+1}}{2}.$
\STATE 6: $z^{k+1}=h^{k+1}+\frac{\alpha_k-1}{\alpha_{k+1}}(h^{k+1}-h_k)$.
\STATE 7: $k\leftarrow k+1$.
\UNTIL\{The stopping criterion \eqref{eqn3-8} is satisfied.\}
\STATE 8: $H_{\cdot j}^{t+1}=z^k$.
\end{algorithmic}
\end{algorithm}

The complete procedure of the HQ algorithm is summarized in \textbf{Algorithm \ref{alg3-2}}. The weights of entries and factor matrices are updated recursively until the objective function does not change. We use the following stopping criterion to check the convergence in \textbf{Algorithm \ref{alg3-2}}:
\begin{equation}
\label{eqn3-9}
\frac{|F(W^t,H^t)-F(W^*,H^*)|}{|F(W^0,H^0)-F(W^t,H^t)|}\le\epsilon_2,
\end{equation}
where $\epsilon_2$ signifies the tolerance, $F(W,H)$ signifies the objective function of \eqref{eqn3-1} and $(W^*,H^*)$ signifies a local minimizer\footnote{Since any local minimal is unknown beforehand, we instead utilize $(W^{t-1},H^{t-1})$ in our experiments.}. The stopping criterion \eqref{eqn3-9} implies that HQ stops when the search point is sufficiently close to the minimizer  and sufficiently far from the initial point. Line $3$ updates the scale parameter by the Nagy algorithm and will be further presented in Section \ref{sec:chapter3-2}. Line $4$ detects outliers by robust statistics and will be presented in Section \ref{sec:chapter3-3}.

\begin{algorithm}
\caption{Half-quadratic (HQ) Programming Algorithm for Truncated CauchyNMF}
\label{alg3-2}
\begin{algorithmic}
\STATE \textbf{Input}: $W\in\mathbb{R}_+^{m\times n}$, $r\ll\min\{m,n\}$.
\STATE \textbf{Output}: $W,H$.
\STATE 1: Initialize $W^0\in\mathbb{R}_+^{m\times r}$, $H^0\in\mathbb{R}_+^{r\times n}$, $t=0$.
\REPEAT
\STATE 2: Calculate $E^t=V-W^t H^t$ and $Q^{t+1}=\frac{1}{(1+(\frac{E^t}{\gamma})^2)}$.
\STATE 3: Update the scale parameter $\gamma$ based on $E^t$.
\STATE 4: Detect the indices $\Omega(t)$ of outliers and set $Q_{\Omega(t)}^{t+1}=0$.
\FOR{$j=1,\dots,n$}
\STATE 5: Calculate $D_j^{t+1}=diag(Q_{\cdot j}^{t+1})$.
\STATE 6: Update $H_{\cdot j}^{t+1}$ by \textbf{Algorithm} \ref{alg3-1}.
\ENDFOR
\STATE 7: Calculate $E^t=V-W^t H^{t+1}$ and $Q^{t+1}=\frac{1}{(1+(\frac{E^t}{\gamma})^2)}$.
\FOR{$i=1,\dots,m$}
\STATE 8: Calculate $D_i^{t+1}=diag(Q_{i\cdot}^{t+1})$.
\STATE 9: Update $W_{i\cdot}^{t+1}$ by \textbf{Algorithm} \ref{alg3-1}.
\ENDFOR
\STATE 10: $t\leftarrow t+1$.
\UNTIL\{Stopping criterion (\ref{eqn3-9}) is satisfied.\}
\STATE 11: $W=W^t, H=H^t$.
\end{algorithmic}
\end{algorithm}

The main time cost of \textbf{Algorithm \ref{alg3-2}} is incurred on lines $2$, $4$, $5$, $6$, $7$, $8$, and $9$. The time complexities of lines $2$ and $7$ are both $O(mnr)$. According to \textbf{Algorithm \ref{alg3-1}}, the time complexities of lines $6$ and $9$ are $O(mr^2)$ and $O(nr^2)$, respectively. Since line $4$ introduces a median operator, its time complexity is $O(mn\ln(mn))$. In summary, the total complexity of \textbf{Algorithm \ref{alg3-2}} is $O((mn\ln(mn)+mnr^2))$.

\subsection{Scale Estimation}
\label{sec:chapter3-2}
The parameter estimation problem for Cauchy distribution has been studied for several decades \cite{bib51}\cite{bib52}\cite{bib53}. Nagy \cite{bib51} proposed an I-divergence based method, termed the Nagy algorithm for short, to simultaneously estimate location and scale parameters. The Nagy algorithm minimizes the discrimination information\footnote{The discrimination information of random variable $\xi_1$ given random variable $\xi_2$ is defined as $D(\xi_1|\xi_2)=\int_{-\infty}^{+\infty}\ln\frac{f_1(x)}{f_2(x)}dF_1(x)$, where $f_1$ and $f_2$ are the PDFs of $\xi_1$ and $\xi_2$, and $F_1$ is the distribution function of $\xi_1$.} between the empirical distribution of the data points and the prior Cauchy distribution with respect to the parameters. In our Truncated CauchyNMF model \eqref{eqn2-4}, the location parameter of the Cauchy distribution is assumed to be zero, and thus we only need to estimate the scale-parameter $\gamma$.

Here we employ the Nagy algorithm to estimate the scale-parameter based on all the residual errors of the data. According to \cite{bib51}, supposing there exist a large number of residual errors, the scale-parameter estimation problem can be formulated as
\begin{eqnarray}
\label{eqn3-10}
&\min_\gamma D(\eta_n|f_{0,\gamma})&=\min_\gamma\int_{\infty}^{+\infty}\ln\frac{1}{f_{0,\gamma}(x)}dF_n(x)\nonumber\\
&&=\min_\gamma\sum_{n=1}^N\frac{1}{N}\ln\frac{1}{f_{0,\gamma}(x_k)},
\end{eqnarray}
where $D(\cdot|\cdot)$ denotes the discrimination information, and the first equality is due to the independence of $\eta_n$ and $\gamma$, and the second equality is due to the Law of large numbers. By substituting the probability density function $f_{0,\gamma}$ of Cauchy distribution\footnote{The probability density function (PDF) of Cauchy distribution is $f_{x_0,\gamma}(x)={1}/({\pi\gamma(1+(\frac{x-x_0}{\gamma})^2)})$, where $x_0$ is the location parameter, specifying the location of the peak of the distribution, and $\gamma$ is the scale parameter, specifying the half-width at half-maximum.} into \eqref{eqn3-10} and replacing $\{x_n\}$ with $\{E_{ij}\}$, we can rewrite (\ref{eqn3-10}) as follows: $\min_\gamma\sum_{i=1}^m\sum_{j=1}^n\frac{1}{mn}\ln\{\pi\gamma(1+(\frac{E_{ij}}{\gamma})^2)\}$. To solve this problem, Nagy \cite{bib51} proposed an efficient iterative algorithm, i.e.,
\begin{eqnarray}
\label{eqn3-11}
\gamma_{k+1}=\gamma_k\sqrt{{1}/{e_k^0}-1},k=0,1,2,\dots,
\end{eqnarray}
where $\gamma_0>0$, and $e_k^0=\frac{1}{mn}\sum_{i=1}^m\sum_{j=1}^n\frac{1}{(1+(\frac{E_{ij}}{\gamma_k})^2)}$. In \cite{bib51}, Nagy proved that the algorithm \eqref{eqn3-11} converges to a fixed point assuming the number of data points is large enough, and this assumption is reasonable in Truncated CauchyNMF.

\subsection{Outlier Rejection}
\label{sec:chapter3-3}
Looking more carefully at \eqref{eqn3-5}, \eqref{eqn3-6} and \eqref{eqn3-7}, HQ intrinsically assigns a weight for each entry of $V$ with both factor matrices $H^{t+1}$ and $W^{t+1}$ fixed, i.e., $Q_{ij}^{t+1}=\left\{\begin{array}{cc}\frac{1}{1+(\frac{E^t}{\gamma})_{ij}^2},&if\;|E^t_{ij}|\le\gamma\sqrt{\sigma}\\0,&if\;|E^t_{ij}|>\gamma\sqrt{\sigma}\end{array}\right.$, where $E^t$ denotes the error matrix at the $t$-th iteration. The larger the magnitude of error for a particular entry, the lighter the weight is assigned to it by HQ. Intuitively, the corrupted entry contributes less in learning the intrinsic subspace. If the magnitude of error exceeds a threshold $\gamma\sqrt{\sigma}$, Truncated CauchyNMF assigns zero weights to the corrupted entries to inhibit their contribution to the learned subspace. That is how Truncated CauchyNMF filters out extreme outliers.

However, it is non-trivial to estimate the threshold $\gamma\sqrt{\sigma}$. Here, we introduce a robust statistics-based method to explicitly detect the support of the outliers instead of estimating the threshold to detect outliers. Since the energy function of Truncated CauchyNMF gets close to that of NMF as the error tends towards zero, i.e., $\lim_{x\rightarrow 0}(\ln(1+x^2)-x^2)=0$. Truncated CauchyNMF encourages the small magnitude errors to have a Gaussian distribution. Let $\Theta^t$ denote the set of magnitudes of error at the $t$-th iteration of HQ, i.e., $\Theta^t=\{|E_{ij}^t|: 1\le i\le m,1\le j\le n\}$ where $E^t=V-W^t H^t$. It is reasonable to believe that a subset of $\Theta^t$, i.e., $\Gamma^t=\{\theta\in\Theta^t: \theta\le med\{\Theta^t\}\}$, obeys a Gaussian distribution, where $med\{\Theta^t\}$ signifies the median of $\Theta^t$. Since $|\Gamma^t|=\lfloor\frac{mn}{2}\rfloor$, it suffices to estimate both the mean $\mu^t$ and standard deviation $\delta^t$ from $\Gamma^t$. According to the three-sigma-rule, we detect the outliers as $O^t=\{\tau\in\Gamma^t:|\tau-\mu^t|>3\delta^t\}$ and output their indices $\Omega(t)$.

\begin{figure}[!b]
\centering
\includegraphics[width=1.0\linewidth]{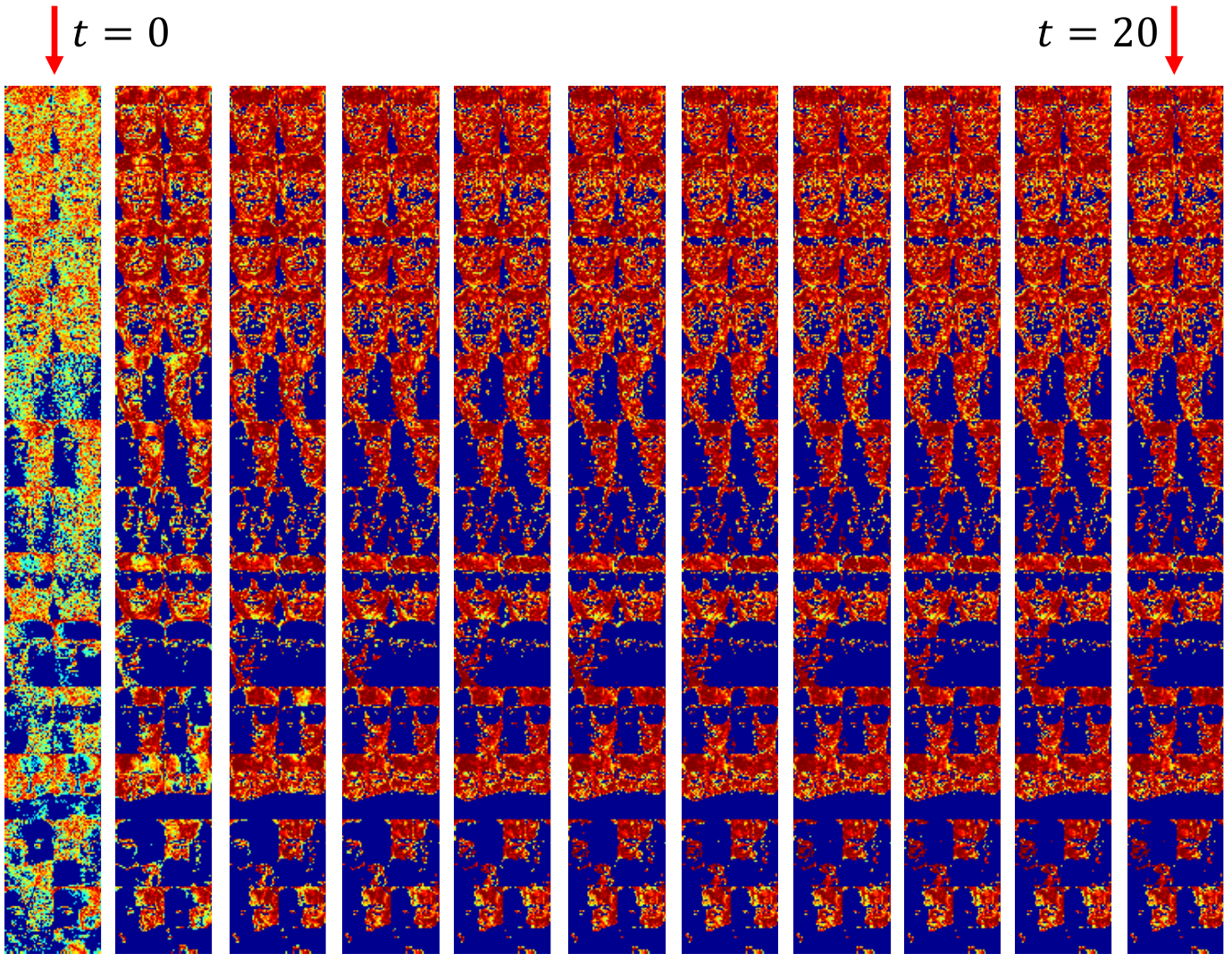}
\caption{An illustrative example of the sequence of weights generated by the HQ algorithm.}
\label{fig3-1}
\end{figure}

To illustrate the effect of outlier rejection, Figure \ref{fig3-1} presents a sequence of weighting matrices generated by HQ for the motivating example described in Figure \ref{fig1-2}. It shows that HQ correctly assigns zero weights for the corrupted entries in only a few iterations and finally detects almost all outliers including illumination, sunglasses, and scarves (see the last column in Figure \ref{fig3-1}) in the end.

\section{Related Work}\label{sec:chapter4}
Before evaluating the effectiveness and robustness of Truncated CauchyNMF, we briefly review the state-of-the-art of non-negative matrix factorization (NMF) and its robustified variants. We have thoroughly compared the robustness between the proposed Truncated CauchyNMF and all the listed related works.

\subsection{NMF}
Traditional NMF \cite{bib27} assumes that noise obeys a Gaussian distribution and derives the following squared $L_2$-norm based objective function: $\min_{W\ge 0,H\ge 0}\|V-WH\|_F^2$, where $\|X\|_F=\sqrt{\sum_{ij}X_{ij}^2}$ signifies the matrix Frobenius norm. It is commonly known that NMF can be solved by using the multiplicative update rule (MUR, \cite{bib27}). Because of the nice mathematical property of squared $L_2$-norm and the efficiency of MUR, NMF has been extended for various applications \cite{bib5}\cite{bib8}\cite{bib44}. However, NMF and its extensions are non-robust because the $L_2$-norm is sensitive to outliers.

\subsection{Hypersurface Cost Based NMF}
\label{sec:chapter4-2}
Hamza and Brady \cite{bib18} proposed a hypersurface cost based NMF (HCNMF) by minimizing the summation of hypersurface costs of errors, i.e., $\min_{W\ge0,H\ge0}\{\sum_{ij}\delta((V-WH)_{ij})\}$, where $\delta(x)=\sqrt{1+x^2}-1$ is the hypersurface cost function. According to \cite{bib18}, the hypersurface cost function has differentiable and bounded influence function. Since the hypersurface cost function is differentiable, HCNMF can be directly solved by using the projected gradient method. However, the optimization of HCNMF is difficult because Armijo's rule based line search is time consuming \cite{bib18}.

\subsection{$L_1$-Norm Based NMF}
To improve the robustness of NMF, Lam \cite{bib25} assumed that noise is independent and identically distributed from Laplace distribution and proposed $L_1$-NMF as follows: $\min_{W\ge0,H\ge0}\|V-WH\|_1$, where $\|X\|_1=\sum_{ij}|X_{ij}|$ and $|\cdot|$ signifies the absolute value function. Since the $L_1$-norm based loss function is non-smooth, the optimization algorithm in \cite{bib25} is not scalable on large-scale datasets. Manhattan NMF (MahNMF, \cite{bib17}) remidies this problem by approximating the loss function of $L_1$-NMF with a smooth function and minimizing the approximated loss function using Nesterov's method. Although $L_1$-NMF is less sensitive to outliers than NMF, it is not sufficiently robust because its breakdown point is related to the dimensionality of data \cite{bib9}.

\subsection{$L_1$-Norm Regularized Robust NMF}
Zhang \textit{et al}. \cite{bib48} assumed that the dataset contains both Laplace distributed noise and Gaussian distributed noise and proposed an $L_1$-norm regularized Robust NMF (RNMF-$L_1$) as follows: $\min_{W\ge0,H\ge0,S}\{\|V-WH-S\|_F^2+\lambda\|S\|_1\}$, where $\lambda$ is a positive constant that trades off the sparsity of $S$. Similar to $L_1$-NMF, RNMF-$L_1$ is also less sensitive to outliers than NMF, but they are both non-robust to large numbers of outliers because the $L_1$-minimization model has a low breakdown point. Moreover, it is non-trivial to determine the tradeoff parameter $\lambda$.

\subsection{$L_{2,1}$-Norm Based NMF}
Since NMF is substantially a summation of the squared $L_2$-norm of the errors, the large magnitude errors dominate the objective function and cause NMF to be non-robust. To solve this problem, Kong \textit{et al}. \cite{bib24} proposed the $L_{2,1}$-norm based NMF ($L_{2,1}$-NMF) which minimizes the $L_{2,1}$-norm of the error matrix, i.e., $\min_{W\ge0,H\ge0}\|V-WH\|_{2,1}$, where the $L_{2,1}$-norm is defined as $\|E\|_{2,1}=\sum_{j=1}^n\|E_{\cdot j}\|_2$. In contrast to NMF, $L_{2,1}$-NMF is more robust because the influences of noisy examples are inhibited in learning the subspace.

\subsection{Robust Capped Norm NMF}
Gao \textit{et al}. \cite{bib70} proposed a robust capped norm NMF (RCNMF) to completely filter out the effect of outliers by instead minimizing the following objective function: $\sum_{W\ge 0,H\ge 0}\sum_{j=1}^n\min\{\|V_{\cdot j}-WH_{\cdot j}\|,\theta\}$, where $\theta$ is a threshold that chooses the outlier samples. RCNMF cannot be applied in practical applications because it is non-trivial to determine the pre-defined threshold, and the utilized iterative algorithms in both \cite{bib24} and \cite{bib70} converge slowly with the successive use of the power method \cite{bib2}.

\subsection{Correntropy Induced Metric Based NMF}
The most closely-related work is the half-quadratic algorithm for optimizing robust NMF, which includes the Correntropy-Induced Metric (CIM)-based NMF (CIM-NMF) and Huber-NMF by Du \textit{et al}. \cite{bib10}. CIM-NMF measures the approximation errors by using CIM \cite{bib30}, i.e., $\min_{W\ge0,H\ge0}\sum_{i=1}^m\sum_{j=1}^n\rho((V-WH)_{ij},\delta)$, where $\rho(x,\delta)=1-\frac{1}{\sqrt{2\pi}\delta} e^{-\frac{x^2}{2\delta^2}}$. Since the energy function $\rho(x,\delta)$ increases slowly as the error increases, CIM-NMF is insensitive to outliers. In a similar way, Huber-NMF \cite{bib10} measures the approximation errors by using the Huber function, i.e., $\min_{W\ge0,H\ge0}\sum_{i=1}^m\sum_{j=1}^n\mathit{l}((V-WH)_{ij},c)$, where $\mathit{l}(x,c)=\left\{\begin{array}{cc}x^2,&|x|\le c\\2c|x|-c^2,&|x|\ge c\end{array}\right.$ and the cutoff $c$ is automatically determined by $c=med\{|(V-WH)_{ij}|\}$.

Truncated CauchyNMF is different from both CIM-NMF and Huber-NMF in four aspects: (1) Truncated CauchyNMF is derived from the proposed Truncated Cauchy loss which can model both modearte and extreme outliers, whereas neither CIM-NMF or Huber-NMF can do that; (2) Truncated CauchyNMF demonstrates strong evidence of both robustness and generalization ability, whereas neither CIM-NMF nor Huber-NMF demonstrates evidence of neither; (3) Truncated CauchyNMF iteratively detects outliers by the robust statistics on the magnitude of errors, and thus performs more robustly than CIM-NMF and Huber-NMF in practice; And (4) Truncated CauchyNMF obtains the optima for each factor in each iteration round by solving the weighted non-negative least squares (WNLS) problems, whereas the multiplicative update rules for CIM-NMF and Huber-NMF do not.

\section{Experimental Verification}\label{sec:chapter5}
We explore both the robustness and the effectiveness of Truncated CauchyNMF on two popular face image datasets, ORL \cite{bib42} and AR \cite{bib33}, and one object image dataset, i.e., Caltech $101$ \cite{bib67}, by comparing with six typical NMF models: (1) $L_2$-NMF \cite{bib26} optimized by NeNMF \cite{bib16}; (2) $L_1$-NMF \cite{bib25} optimized by MahNMF \cite{bib17}; (3) RNMF-$L_1$ \cite{bib48}; (4) $L_{2,1}$-NMF \cite{bib24}; (5) CIM-NMF \cite{bib10}; and (6) Huber-NMF \cite{bib10}. We first present a toy example to intuitively show the robustness of Truncated CauchyNMF and several clustering experiments on the contaminated ORL dataset to confirm its robustness. We then analyze the effectiveness of Truncated CauchyNMF by clustering and recognizing face images in the AR dataset, and clustering object images in the Caltech $101$ dataset.

\begin{figure}[!t]
\centering
\includegraphics[width=1.0\linewidth]{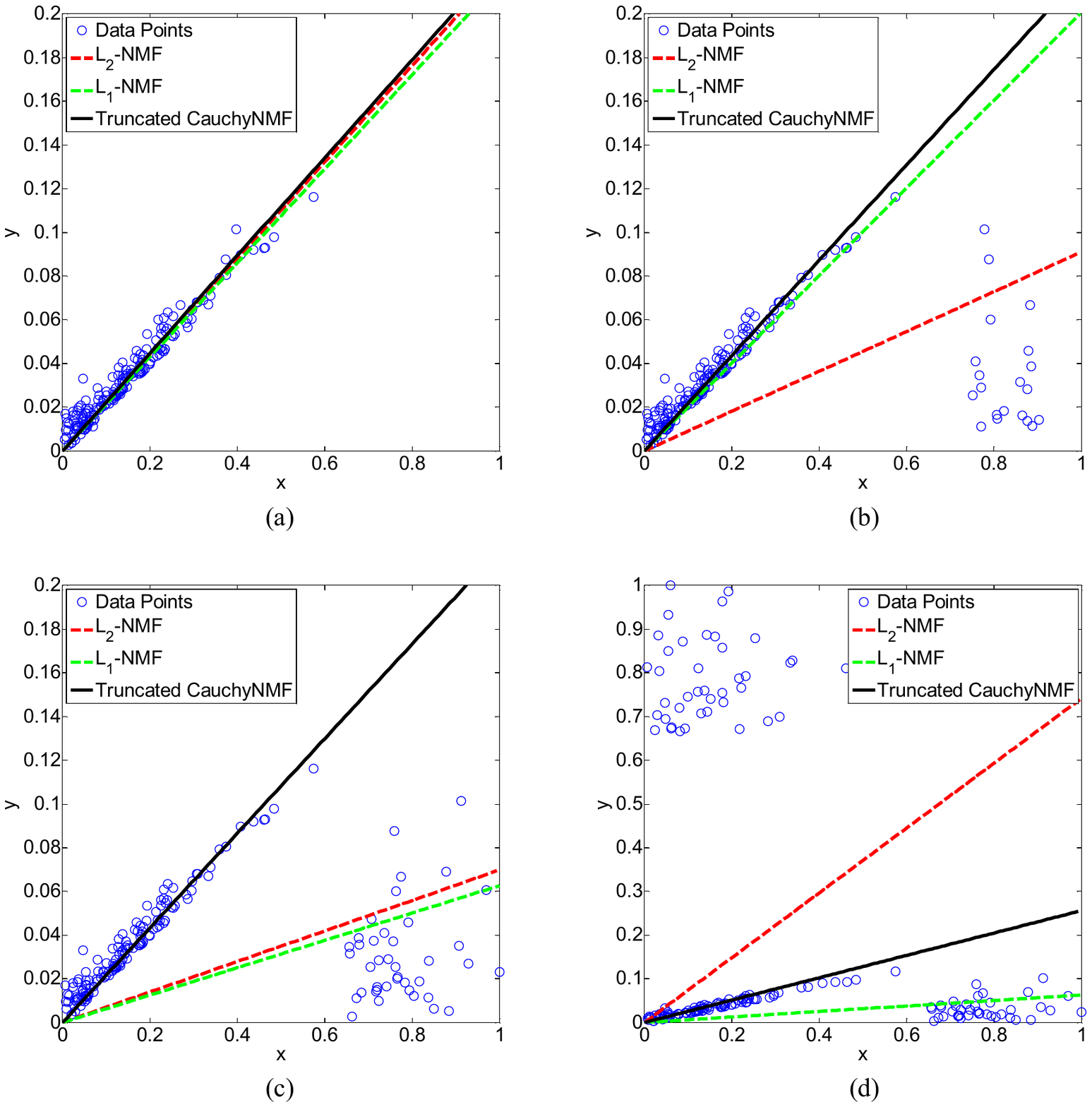}
\caption{Learning the one-dimensional subspace (i.e., a straight line) from $180$ synthetic two-dimensional data points by $L_2$-NMF, $L_1$-NMF, and Truncated CauchyNMF in four cases: (a) clean dataset, (b) $20$ points contaminated in $x$-direction, (c) $40$ points contaminated in $x$-direction, and (d) $80$ points contaminated in both directions.}
\label{fig5-1}
\end{figure}

\subsection{An Illustrative Study}
\label{sec:chapter5-1}
To illustrate Truncated CauchyNMF's ability to learn a subspace, we apply Truncated CauchyNMF on a synthetic dataset composed of $180$ two-dimensional data points (see Figure \ref{fig5-1}(a)). All data points are distributed in a one-dimensional subspace, i.e., a straight line ($y=0.2x$). Both $L_2$-NMF and $L_1$-NMF are applied on this synthetic dataset for comparison.

Figure \ref{fig5-1}(a) shows that all methods learn the intrinsic subspace correctly on the clean dataset. Figures \ref{fig5-1}(b) to \ref{fig5-1}(d) demonstrate the robustness of Truncated CauchyNMF on a noisy dataset. First, we randomly select $20$ data points and contaminate their $x$-coordinates, with their y-coordinates retained to simulate outliers. Figure \ref{fig5-1}(b) shows that $L_2$-NMF fails to recover the subspace in the presence of $\frac{1}{9}$ outliers, while both Truncated CauchyNMF and $L_1$-NMF perform robustly in this case. However, the robustness of $L_1$-NMF decreases as the outliers increase. To study this point, we randomly select another $20$ data points and contaminate their $x$-coordinates. Figure \ref{fig5-1}(c) shows that both $L_2$-NMF and $L_1$-NMF fail to recover the subspace, but Truncated CauchyNMF succeeds. To study the robustness of Truncated CauchyNMF on seriously corrupted datasets, we randomly select an additional $40$ data points as outliers. We contaminate their $y$-coordinates while keeping their $x$-coordinates consistent. Figure \ref{fig5-1}(d) shows that Truncated CauchyNMF still recovers the intrinsic subspace in the presence of $\frac{4}{9}$ outliers while both $L_2$-NMF and $L_1$-NMF fail in this case. In other words, the breakdown point of Truncated CauchyNMF is greater than $44.4\%$, which is quite close to the highest breakdown point of $50\%$.

\begin{table*}[!t]
\caption{Relative reconstruction error (\%) of $L_2$-NMF, $L_{2,1}$-NMF, RNMF-$L_1$, $L_1$-NMF, Huber-NMF, CIM-NMF, and CauchyNMF on ORL dataset contaminated by Laplace noise with deviation varying from $40$ to $280$.}
\label{tbl5-1}
\centering
\begin{tabular}{c||ccccccc}
\hline
$\delta$  &$L_2$-NMF  &$L_{2,1}$-NMF  &RNMF-$L_1$ &$L_1$-NMF  &Huber-NMF  &CIM-NMF  &Truncated CauchyNMF\\
\hline
40  &14.78(0.01)  &16.68(0.04)  &17.18(0.09)  &13.56(0.04)  &14.05(0.07)  &15.93(0.08)  &13.41(0.04)\\
80  &24.91(0.02)  &25.39(0.03)  &21.30(0.11)  &17.18(0.05)  &17.53(0.04)  &16.27(0.06)  &14.70(0.06)\\
120 &36.30(0.02)  &35.65(0.07)  &24.66(0.08)  &21.33(0.06)  &21.87(0.07)  &18.95(0.05)  &15.94(0.07)\\
160 &47.48(0.03)  &46.08(0.04)  &27.49(0.06)  &25.38(0.06)  &26.47(0.08)  &22.11(0.05)  &16.88(0.13)\\
200 &59.18(0.04)  &57.35(0.04)  &30.27(0.11)  &29.73(0.10)  &31.72(0.16)  &25.84(0.07)  &18.10(0.13)\\
240 &70.70(0.03)  &68.52(0.07)  &32.96(0.15)  &33.98(0.17)  &37.12(0.55)  &29.68(0.11)  &19.88(0.55)\\
280 &82.06(0.04)  &79.78(0.09)  &35.81(0.17)  &38.13(0.37)  &43.07(0.62)  &33.67(0.12)  &27.23(4.06)\\
\hline
\end{tabular}
\end{table*}

\subsection{Simulated Corruption}
\begin{figure}[!t]
\centering
\includegraphics[width=0.6\linewidth]{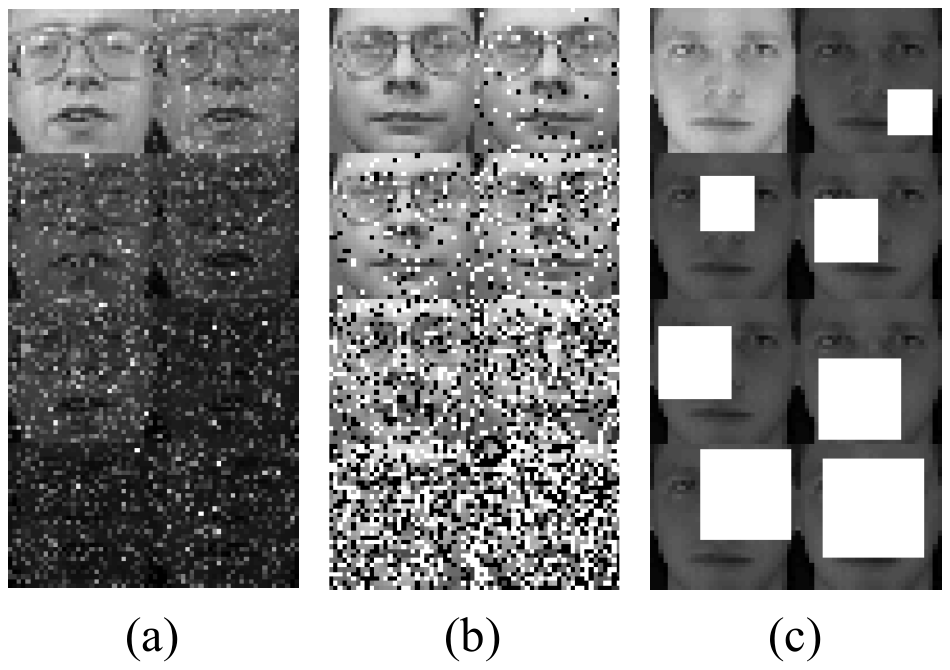}
\caption{Example face images of ORL database: (a) an example face image and its noised versions by Laplace noise with deviation $\delta=40,80,120,160,200,240,280$, (b) an example face image and its noisy versions, where $p\%$ pixels are contaminated by Salt \& Pepper noise and $p=5,10,20,30,40,50,60$, (c) an example face image and its occluded versions by $b\times b$-blocks with $b=10,12,14,16,18,20,22$.}
\label{fig5-2}
\end{figure}

We first evaluate Truncated CauchyNMF' robustness to simulated corruptions. To this end, we add three typical corruptions, i.e., Laplace noise, and Salt \& Pepper noise, randomly positioned blocks, to frontal face images from the Cambridge ORL database and compare the clustering performance of our methods with the performance of other methods on these contaminated images. Figure \ref{fig5-2} shows example face images contaminated by these corruptions.

The Cambridge ORL database \cite{bib42} contains $400$ frontal face photos of $40$ individuals. There are $10$ photos of each individual with a variety of lighting, facial expressions and facial details (with-glasses or without-glasses). All photos were taken against the same dark background and each photo was cropped to a $32\times32$ pixel array and normalized to a long vector. The clustering performance is evaluated by two metrics, namely accuracy and normalized mutual information \cite{bib33}. The number of clusters is set equal to the number of individuals, i.e., $40$. Intuitively, the better a model clusters contaminated images, the more robust it is for learning the subspace. In this experiment, we utilize K-means \cite{bib32} as a baseline. To qualify the robustness of all NMF models, we compare their relative reconstruction errors, i.e., ${\|\hat{V}-WH\|_F}/{\|\hat{V}\|_F}$, where $\hat{V}$ denotes the clean dataset, and $W$ and $H$ signify the factorization results on the contaminated dataset.

\subsubsection{Laplace Noise}
\begin{figure}[!t]
\centering
\includegraphics[width=1.0\linewidth]{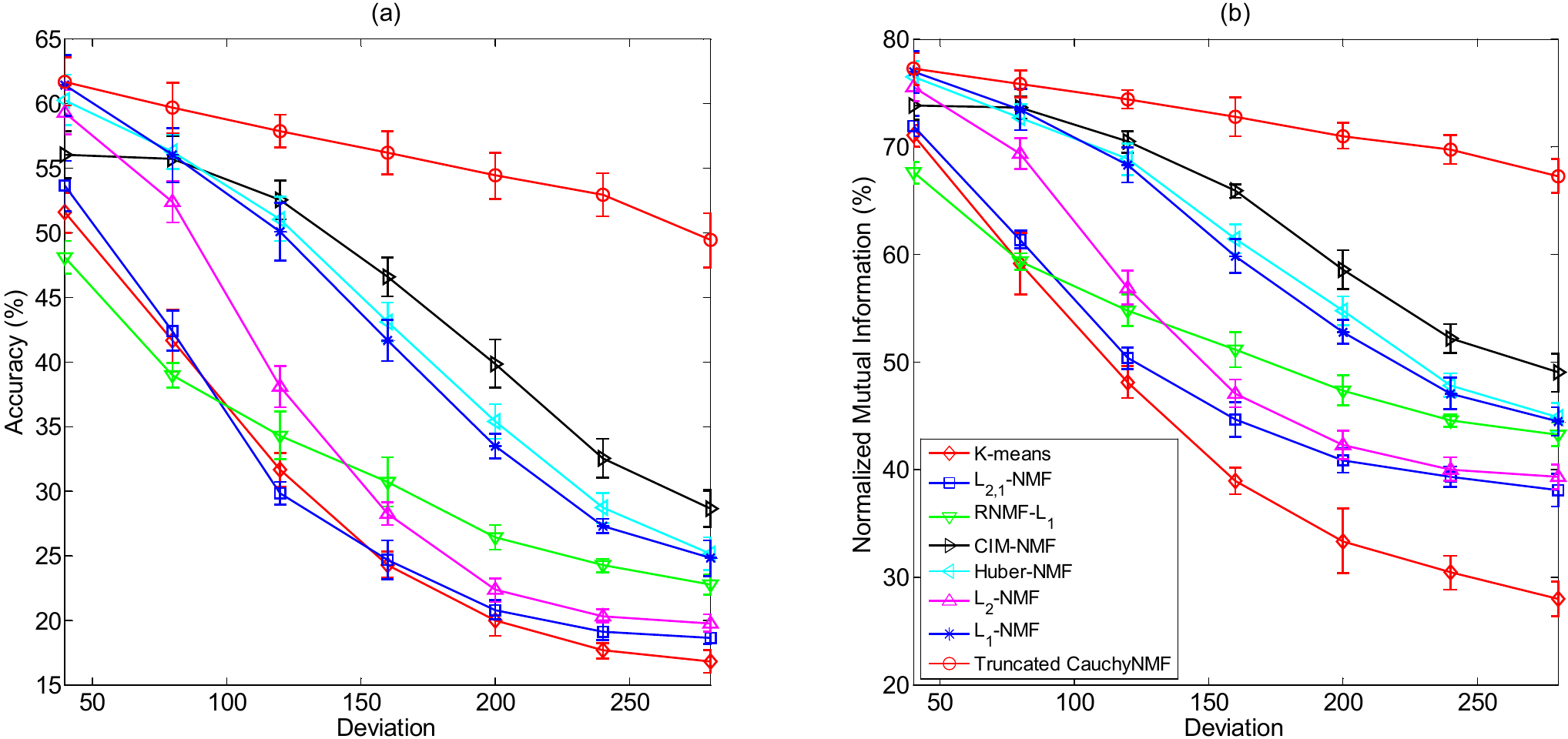}
\caption{Evaluation on frontal face images of ORL database contaminated by Laplace noise: (a) average accuracy and standard deviation of K-means, $L_2$-NMF, $L_{2,1}$-NMF, RNMF-$L_1$, $L_1$-NMF, Huber-NMF, CIM-NMF and Truncated CauchyNMF, (b) comparison of average normalized mutual information and standard deviation.}
\label{fig5-3}
\end{figure}

Laplace noise exists in many types of observation, e.g., gradient-based image features such as SIFT \cite{bib50}, but the classical NMF cannot deal with such data because the distributions violate the assumption of classical NMF. In this experiment, we study Truncated CauchyNMF's capacity to deal with Laplace noisy data. We simulate Laplace noise by adding random noise to each pixel of each face image from ORL where the noise obeys a Laplace distribution $Laplace(0,\delta)$. For the purpose of verifying the robustness of Truncated CauchyNMF, we vary the deviation $\delta$ from $40$ to $280$ because the maximum pixel value is $255$. Figure \ref{fig5-2}(a) gives an example face image and its seven noisy versions by adding Laplace noise. Figure \ref{fig5-3}(a) and \ref{fig5-3}(b) present the mean and standard deviations of accuracy and normalized mutual information of Truncated CauchyNMF and the representative models.

Figure \ref{fig5-3} confirms that NMF models outperform K-means in terms of accuracy and normalized mutual information. $L_1$-NMF outperforms $L_2$-NMF and $L_{2,1}$-NMF because $L_1$-NMF models Laplace noise better. $L_1$-NMF outperforms RNMF-$L_1$ because $L_1$-NMF assigns smaller weight for large noise than RNMF-$L_1$. CIM-NMF and Huber-NMF perform comparably with $L_1$-NMF when the deviation of Laplace noise is moderate. However, as the deviation increases, their performance is dramatically reduced because large-magnitude outliers seriously influence the factorization results. In contrast, Truncated CauchyNMF outperforms all the representative NMF models and remains stable as deviation varies.

The clustering performance in Figure \ref{fig5-3} confirms Truncated CauchyNMF's effectiveness in learning the subspace on the ORL dataset contaminated by Laplace noise. Table \ref{tbl5-1} compares the relative reconstruction errors of Truncated CauchyNMF and the representative algorithms. It shows that CauchyNMF performs the most robustly in all situations. That is because Truncated CauchyNMF can not only model the simulated Laplace noise but also models the underlying outliers, e.g., glasses, in the ORL dataset.

\begin{table*}[!t]
\caption{Relative reconstruction error (\%) of $L_2$-NMF, $L_{2,1}$-NMF, RNMF-$L_1$, $L_1$-NMF, Huber-NMF, CIM-NMF, and Truncated CauchyNMF on ORL dataset contaminated by Salt \& Pepper noise with the percentage of corrupted pixels varying from $5\%$ to $60\%$.
}
\label{tbl5-2}
\centering
\begin{tabular}{c||ccccccc}
\hline
p &$L_2$-NMF  &$L_{2,1}$-NMF  &RNMF-$L_1$ &$L_1$-NMF  &Huber-NMF  &CIM-NMF  &Truncated CauchyNMF\\
\hline
5   &12.51(0.03)  &14.50(0.05)  &14.36(0.10)  &11.33(0.04)  &12.00(0.06)  &13.05(0.09)  &12.37(0.05)\\
10  &15.36(0.02)  &16.44(0.04)  &14.93(0.12)  &11.50(0.05)  &12.03(0.07)  &12.25(0.11)  &12.27(0.06)\\
20  &20.30(0.03)  &19.99(0.07)  &15.97(0.08)  &11.98(0.04)  &12.29(0.04)  &12.18(0.08)  &12.00(0.06)\\
30  &24.44(0.03)  &23.33(0.08)  &17.47(0.11)  &13.25(0.09)  &13.24(0.06)  &13.86(0.09)  &11.80(0.04)\\
40  &28.30(0.03)  &26.59(0.06)  &19.11(0.06)  &16.75(0.08)  &17.23(0.11)  &18.94(0.13)  &12.35(0.06)\\
50  &31.51(0.04)  &29.46(0.06)  &21.71(0.11)  &22.49(0.67)  &26.82(0.33)  &28.54(0.22)  &22.97(0.28)\\
60  &24.28(0.03)  &31.92(0.04)  &26.33(0.13)  &29.62(0.17)  &34.30(0.22)  &39.10(0.63)  &35.26(0.13)\\
\hline
\end{tabular}
\end{table*}

\subsubsection{Salt \& Pepper Noise}
Salt \& Pepper noise is a common type of corruption in images. The removal of Salt \& Pepper noise is a challenging task in computer vision since this type of noise contaminates each pixel by zero or the maximum pixel value, and the noise distribution violates the noise assumption of traditional learning models. In this experiment, we verify Truncated CauchyNMF's capacity to handle Salt \& Pepper noises. We add Salt \& Pepper noise to each frontal face image of the ORL dataset (see Figure \ref{fig5-2}(b) for the contaminated face images of a certain individual) and compare the clustering performance of Truncated CauchyNMF on the contaminated dataset with that of the representative algorithms. To demonstrate the robustness of Truncated CauchyNMF, we vary the percentage of corrupted pixels from $5\%$ to $60\%$. For each case of additive Salt \& Pepper noise, we repeat the clustering test 10 times and report the average accuracy and average normalized mutual information to eliminate the effect of initial points.

\begin{figure}[!t]
\centering
\includegraphics[width=1.0\linewidth]{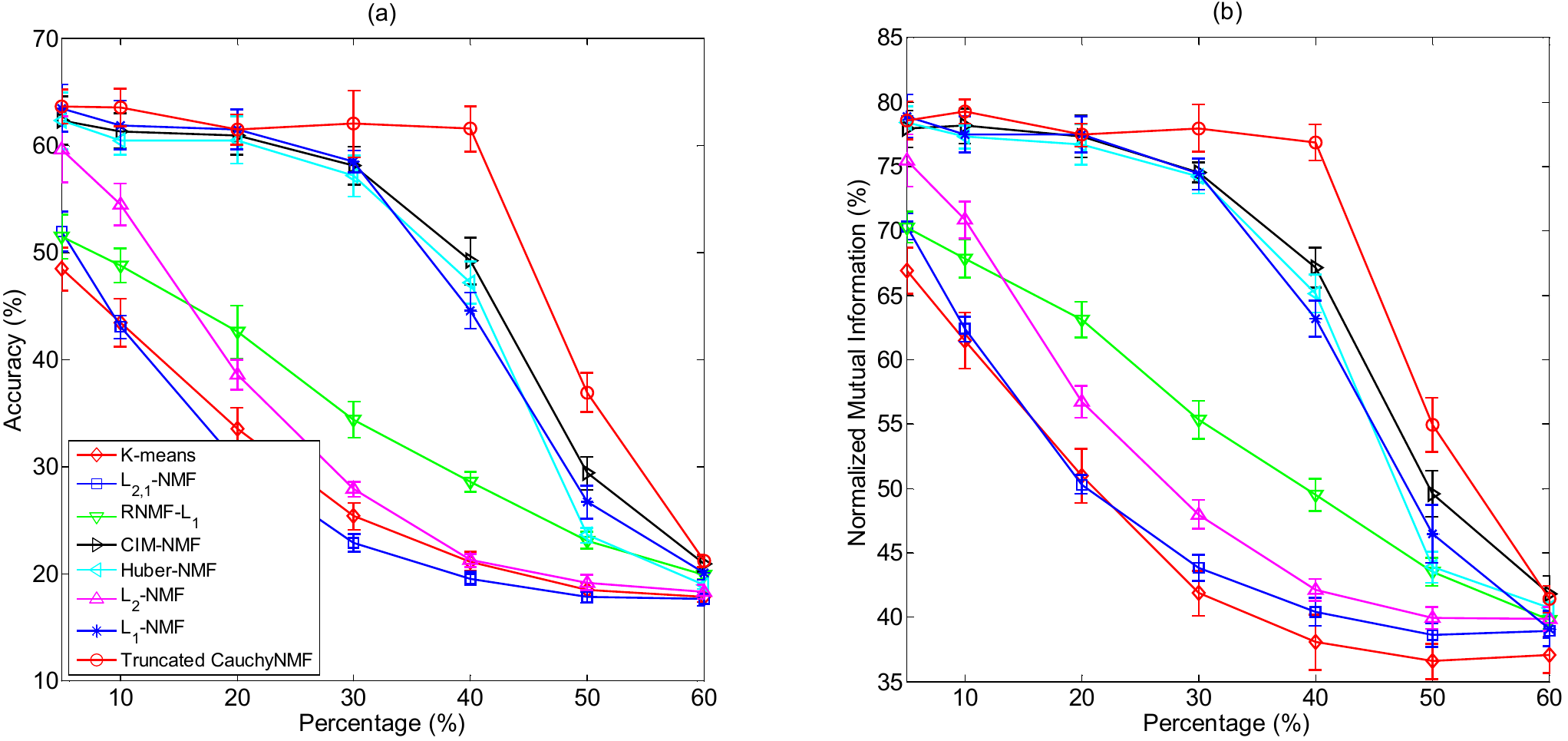}
\caption{Evaluation on frontal face images of ORL database contaminated by Salt \& Pepper noise: (a) average accuracy and standard deviation of K-means, $L_2$-NMF, $L_{2,1}$-NMF, RNMF-$L_1$, $L_1$-NMF, Huber-NMF, CIM-NMF and Truncated CauchyNMF, (b) comparison of average normalized mutual information and standard deviation.}
\label{fig5-4}
\end{figure}

Figure \ref{fig5-4} shows that all models perform satisfactorily when $5\%$ of the pixels of each image are corrupted. As the number of corrupted pixels increases, the classical $L_2$-NMF is seriously influenced by the Salt \& Pepper noise and its performance is dramatically reduced. Although $L_1$-NMF, Huber-NMF and CIM-NMF perform more robustly than $L_2$-NMF, their performance is also degraded when more than $40\%$ of pixels are corrupted. Truncated CauchyNMF performs quite stably even when $40\%$ of pixels are corrupted and outperforms all the representative models in most cases. All the models fail when $60\%$ of pixels are corrupted, because it is difficult to distinguish inliers from outliers in this case.

Table \ref{tbl5-2} gives a comparison of Truncated CauchyNMF and the representative algorithms in terms of relative reconstruction error. It shows that $L_1$-NMF, Hubel-NMF, CIM-NMF and Truncated CauchyNMF perform comparably when less than $20\%$ of the pixels are corrupted, but the robustness of $L_1$-NMF, Hubel-NMF, CIM-NMF are unstable as the percentage of corrupted pixels increases. Truncated CauchyNMF performs stably when $30\%\sim50\%$ of the pixels are corrupted by Salt \& Pepper noise. This confirms the robustness of Truncated CauchyNMF.

\begin{table*}[!t]
\caption{Average accuracy (\%) and average normalized mutual information (\%) of K-means, $L_2$-NMF, $L_{2,1}$-NMF, RNMF-$L_1$, $L_1$-NMF, Huber-NMF, CauchyNMF, CIM-NMF, and Truncated CauchyNMF on occluded ORL dataset with block size $b$ varying from $10$ to $22$ with step size $2$.}
\label{tbl5-3}
\centering
\begin{tabular}{@{\;}c||c@{\quad}c@{\quad}c@{\quad}c@{\quad}c@{\quad}c@{\quad}c@{\quad}c@{\quad}c@{\;}c@{\;}}
\hline
b &K-means  &$L_2$-NMF  &$L_{2,1}$-NMF  &RNMF-$L_1$ &$L_1$-NMF  &Huber-NMF  &CauchyNMF &CIM-NMF &Truncated CauchyNMF\\
\hline
10  &17.20(39.77) &17.10(39.80) &17.20(39.71) &17.50(39.47) &17.65(38.87) &17.68(38.95) &19.27(39.27) &58.48(75.41) &57.80(73.94)\\
12  &17.65(39.95) &17.38(39.43) &17.10(39.90) &17.33(39.46) &17.33(38.79) &17.73(38.96) &18.57(38.96) &56.05(73.36) &58.23(74.31)\\
14  &17.63(40.10) &17.25(39.43) &17.45(39.92) &17.35(39.11) &17.70(39.18) &17.65(38.77) &19.03(39.18) &26.88(46.63) &55.38(71.94)\\
16  &17.55(39.95) &17.25(39.72) &17.20(39.82) &17.25(39.37) &17.43(38.90) &17.35(38.80) &18.36(19.01) &21.18(42.40) &47.30(65.39)\\
18  &16.78(39.09) &17.00(39.06) &16.90(39.73) &16.95(38.67) &16.88(38.20) &17.00(38.41) &17.90(38.48) &23.93(45.21) &42.93(61.84)\\
20  &17.35(39.40) &17.15(39.25) &17.20(39.56) &17.15(38.59) &17.08(38.52) &17.00(38.45) &17.40(38.08) &22.33(43.64) &37.48(57.57)\\
22  &17.15(39.38) &16.75(38.82) &16.88(39.14) &16.90(38.45) &16.95(38.59) &17.10(38.67) &17.73(38.69) &25.38(46.39) &30.05(50.98)\\
\hline
\end{tabular}
\end{table*}

\subsubsection{Contiguous Occlusion}
\label{sec:chapter5-2-3}
The removal of contiguous segments of an object due to occlusion is a challenging problem in computer vision. Many techniques such as $L_1$-norm minimization and nuclear norm minimization are unable to handle this problem. In this experiment, we utilize contiguous occlusion to simulate extreme outliers. Specifically, we randomly position a $b\times b$-sized block on each face image of the ORL dataset and fill each block with a pixel array whose pixel values equal $550$. To verify the effectiveness of subspace learning, we apply both K-means and all NMF models to the contaminated dataset and compare the clustering performance in terms of both accuracy and normalized mutual information. This task is quite challenging because large numbers of outliers with large magnitudes must be ignored to learn a clean subspace. To study the influence of outliers, we vary the block size $b$ from $10$ to $22$, where the minimum block size and maximum block size imply $10\%$ and $50\%$ outliers, respectively. Figure \ref{fig5-2}(c) shows the occluded face images of a certain individual.

\begin{figure}[!t]
\centering
\includegraphics[width=1.0\linewidth]{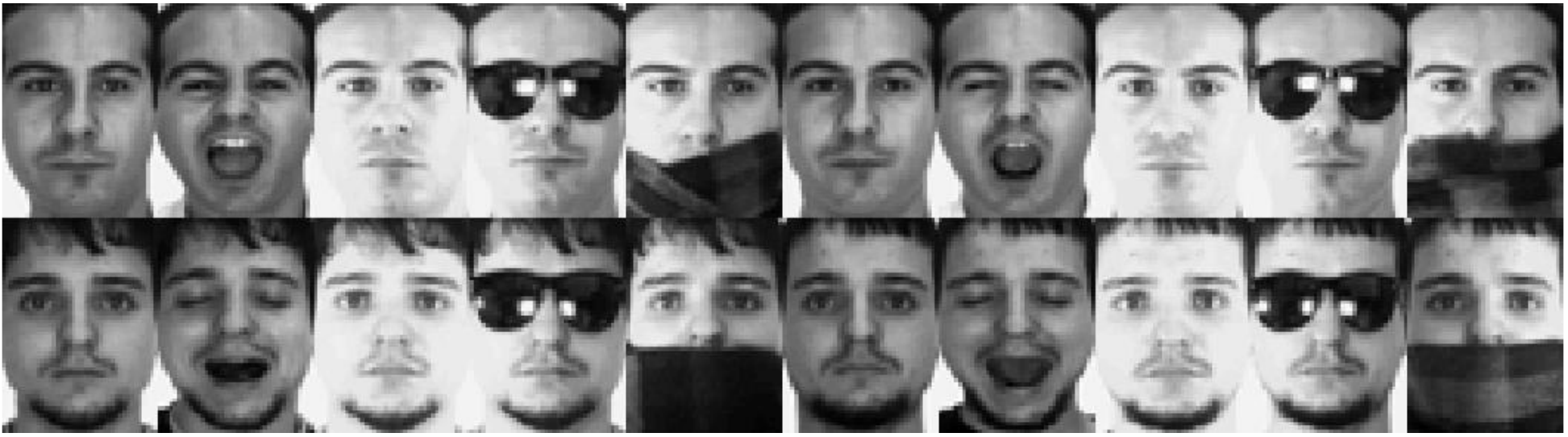}
\caption{Face image examples of two individuals in the AR dataset, with $10$ images per individual.}
\label{fig5-7}
\end{figure}

Table \ref{tbl5-3} shows that K-means, $L_2$-NMF, $L_{2,1}$-NMF, RNMF-$L_1$, $L_1$-NMF, Huber-NMF, and CauchyNMF\footnote{In this experiment, we compare with CauchyNMF to show the effect ot truncation. For CauchyNMF, we set $\sigma=+\infty$ and adopt the proposed HQ algorithm to solve it.} are seriously deteriorated by the added continuous occlusions. Although CIM-NMF performs robustly when the percentage of outliers is moderate, i.e., $10\%$ (corresponds to $b=10) $and $14\%$ (corresponds to $b=12)$, its performance is unstable when the percentage of outliers reaches $20\%$ (corresponds to $b=14$). This is because CIM-NMF keeps energies for extreme outliers and makes a large number of extreme outliers dominate the objective function. By contrast, Truncated CauchyNMF reduces energies of extreme outliers to zeros, and thus performs robustly when the percentage of outliers is less than $40\%$ (corresponds to $b=20)$.



\subsection{Real-life Corruption}
The previous section has evaluated the robustness of Truncated CauchyNMF under several types of synthetic outliers including Laplace noise, Salt \& Pepper noise, and contiguous occlusion. The experimental results show that our methods consistently learns the subspace even when half the pixels in each image are corrupted, while other NMF models fail under this extreme condition. In this section, we evaluate Truncated CauchyNMF's ability to learn the subspace under natural sources of corruption, e.g., contiguous disguise in the AR dataset and object variations in the Caltech-$101$ dataset.

\begin{figure}[!t]
\centering
\includegraphics[width=1.0\linewidth]{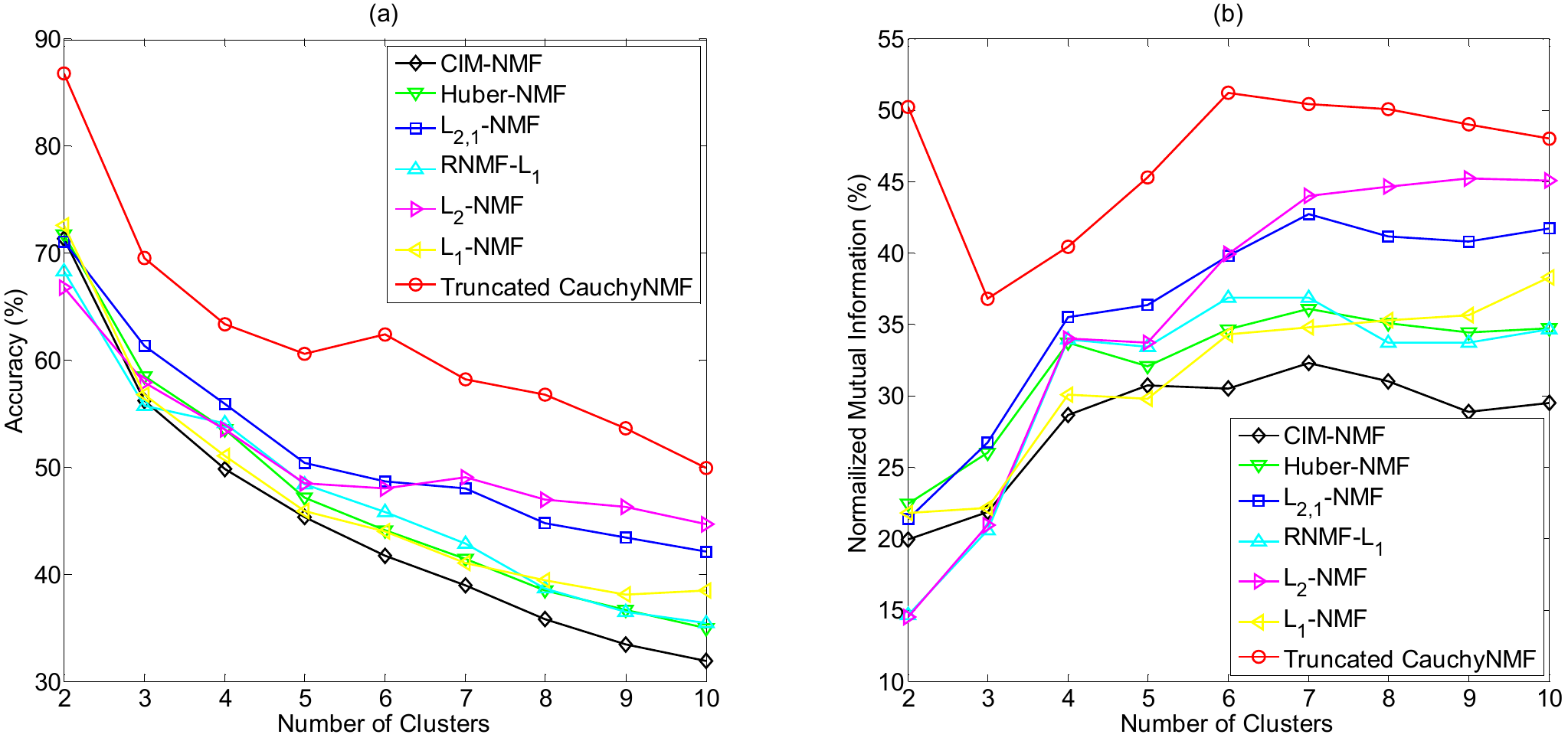}
\caption{Clustering performance in terms of average accuracy and average normalized mutual information of Truncated CauchyNMF, CIM-NMF, Huber-NMF, $L_{2,1}$-NMF, RNMF-$L_1$, $L_2$-NMF, and $L_1$-NMF on the AR dataset, with the number of clusters varying between $2$ and $10$: (a) average accuracy versus number of clusters, and (b) average normalized mutual information versus number of clusters.}
\label{fig5-8}
\end{figure}

\subsubsection{Contiguous Disguise}
\label{sec:chapter5-3-1}
The Purdue AR dataset \cite{bib33} contains $2600$ frontal face images taken from $100$ individuals comprising $50$ males and $50$ females in two sessions.  There is a total of $13$ images in each session, including one normal image, three images depicting different facial expressions, three images under varying illumination conditions, three images with sunglasses, and three images with a scarf for each individual. Each image is cropped into a $55\times40$-dimensional pixel array and reshaped into a $2200$-dimensional long vector. Figure \ref{fig5-7} gives $20$ example images of two individuals and shows that the images with disguises, i.e., sunglasses and scarf, are seriously contaminated by outliers. Therefore, it is quite challenging to correctly group these contaminated images, e.g., the $4$th, $5$th, $9$th and $10$th columns in Figure \ref{fig5-7}, with the clean images, e.g., the $1$st and $6$th columns in Figure \ref{fig5-7}. According to the results in Section \ref{sec:chapter5-2-3}, Truncated CauchyNMF can handle contiguous occlusions with extreme outliers well, we will therefore show the effectiveness of Truncated CauchyNMF to do this job.

To evaluate the effectiveness of Truncated CauchyNMF in clustering, we randomly select between two and ten images of each individual to comprise the dataset. By concatenating all the long vectors, we obtain an image intensity matrix denoted as $V$. We then apply NMF to $V$ to learn the subspace, i.e., $V\approx WH$, where the rank of $W$ and $H$ equals the number of clusters. Lastly, we output the cluster labels by performing K-means on $H$. To eliminate the influence of randomness, we repeat this trial $50$ times and report the averaged accuracy and averaged normalized mutual information for comparison.

Figure \ref{fig5-8} gives both average accuracy and average normalized mutual information in relation to the number of clusters of Truncated CauchyNMF and other NMF models. It shows that Truncated CauchyNMF consistently achieves the highest clustering performance on the AR dataset. This result confirms that Truncated CauchyNMF learns the subspace more effectively than other NMF models, even when the images are contaminated by contiguous disguises such as sunglasses and a scarf.



\begin{table*}[!t]
\caption{Face recognition accuracies (\%) of SEC and NMF models on the AR dataset, with the reduced dimensionalities of NMF models set to $200$ and the test images classified by SRC in the subspaces learned by NMF methods.}
\label{tbl5-4}
\centering
\begin{tabular}{c||cccccc}
\hline
Methods &Total  &Normal &Expressions  &Illuminations  &Scarves  &Sunglasses\\
\hline
Truncated CauchyNMF+SRC &$\mathbf{90}$ &$\mathbf{99}$ &$\mathbf{96.67}$ &92.67 &$\mathbf{90.67}$ &$\mathbf{77}$ \\
CIM-NMF+SRC     &82.54 &96 &92.33 &90.67 &82.33 &60.33 \\
$L_1$-NMF+SRC &87.15 &94 &90 &$\mathbf{95.33}$ &84.67 &76.33\\
$L_2$-NMF+SRC &80.38 &90 &85 &95.33 &79.33 &58.67\\
Huber-NMF+SRC &48.85 &70 &59.33 &51.33 &48.33 &29.33\\
$L_{2,1}$-NMF+SRC &75.23 &87 &83.33 &88 &72 &53.67\\
RNMF-$L_1$+SRC  &49.92 &67 &55.33 &55.33 &52 &31.33\\
SEC         &78.92 &95 &93.33 &90.67 &74.67 &51.67\\
\hline
\end{tabular}
\end{table*}

We further conduct the face recognition experiment on the AR dataset to evaluate the effectiveness of Truncated CauchyNMF. In this experiment, we treat the images taken in the first session as the training set and the images taken in the second session as the test set. This task is challenging because (1) the distribution of the training set is different from that of the test set, and (2) both training and test sets are seriously contaminated by outliers. We first learn a subspace by conducting Truncated CauchyNMF on the whole dataset and then classify each test image by the sparse representation classification method (SRC) \cite{bib69} on the coefficients of both training images and test images in the learned subspace. Since there are totally $100$ individuals and the images of each individual were taken in two sessions, we set the reduced dimensionality of Truncated CauchyNMF to $200$. We also conduct other NMF variants with the same setting for comparison. To filter out the influence of continuous occlusions in face recognition, Zhou \emph{et al}. \cite{bib68} proposed a sparse error correction method (SEC) which labels each pixel of test image as occluded pixel and non-occluded one by using Markov random field (MRF) and learns a representation of each test image on non-occluded pixels. Although SEC succeeds to filter out the continuous occlusions in the test set, it cannot handle outliers in the training set. By contrast, Truncated CauchyNMF can take the occlusions off on both training and test images, and thus boost the performance of the subsequent classification.

Table \ref{tbl5-4} shows the face recognition accuracies of NMF variants and SEC. In the AR dataset, each individual contains one normal image and twelve contaminated images under different conditions including varying facial expressions, illuminations, wearing sunglasses, and wearing scarves. In this experiment, we not only show the results on total test set but also show the results on the test images taken under different conditions separately. Table \ref{tbl5-4} shows that Truncated CauchyNMF performs the best in most cases, especially, it performs almost perfectly on normal images. It validates that Truncated CauchyNMF can learn an effective subspace from the contaminated data. In most situations, SEC performs excellently, but the last two columns indicate that the contaminated training images seriously weaken SEC. Truncated CauchyNMF performs well in such situations because it effectively removes the influence of outliers in the subspace learning stage.

\subsubsection{Object Variation}
\begin{figure}[!t]
\centering
\includegraphics[width=1.0\linewidth]{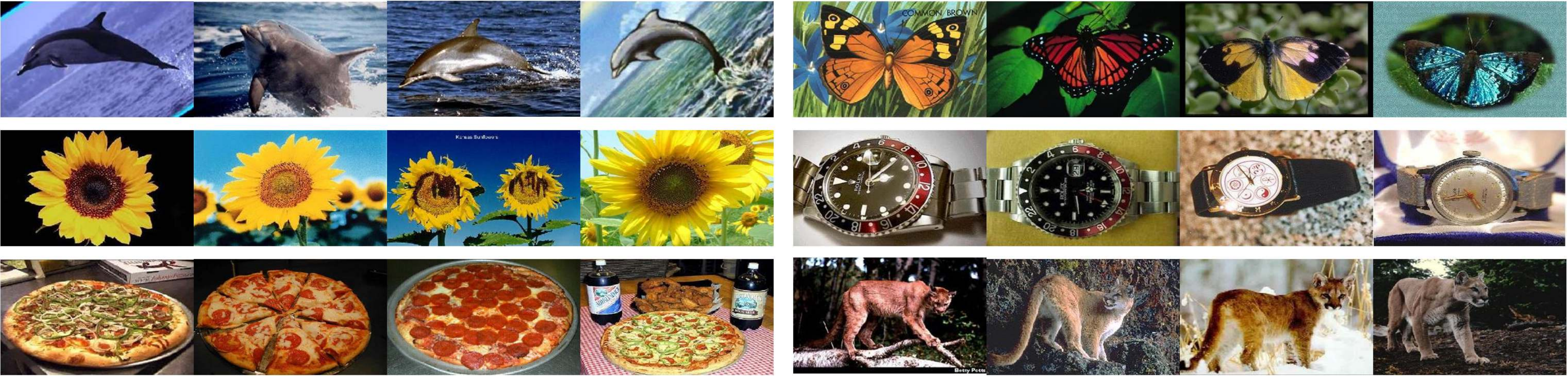}
\caption{Example images in Caltech101 dataset from $6$ categories, and we have four images per category.}
\label{fig5-10}
\end{figure}

The Caltech $101$ dataset \cite{bib67} contains pictures of objects captured from $101$ categories. The number of pictures for each category varies from $40$ to $800$. Figure \ref{fig5-10} shows example images from $6$ different categories including dolphin, butterfly, sunflower, watch, pizza and cougar\_body. We extract convolutional neural network (CNN) feature for each image using the Caffe framework \cite{bib61} and pre-trained model of Imagenet with AlexNet \cite{bib65}. As objects from the same categories may vary in shape, color and size, and the pictures are taken from different viewpoints, clustering objects of the same category together is a very challenging task. We will show the good performance of Truncated CauchyNMF compared to other methods such as CIM-NMF, Huber-NMF, $L_{2,1}$-NMF, RNMF-$L_1$, $L_2$-NMF, $L_1$-NMF, and K-means.

Following the similar protocol as in section \ref{sec:chapter5-3-1}, we demonstrate the effectiveness of Truncated CauchyNMF in clustering objects. We test with $2$ to $10$ randomly selected categories. The image feature matrix is denoted as $V$. NMFs are applied to $V$ to compute the subspace, i.e. $V\approx WH$, where the rank of $W$ and $H$ equals the number of clusters. Cluster labels are obtained by performing K-means on $H$. We repeated such trial $50$ times and computed averaged accuracies and normalized mutual information among all trials for comparison.

\begin{figure}[!t]
\centering
\includegraphics[width=1.0\linewidth]{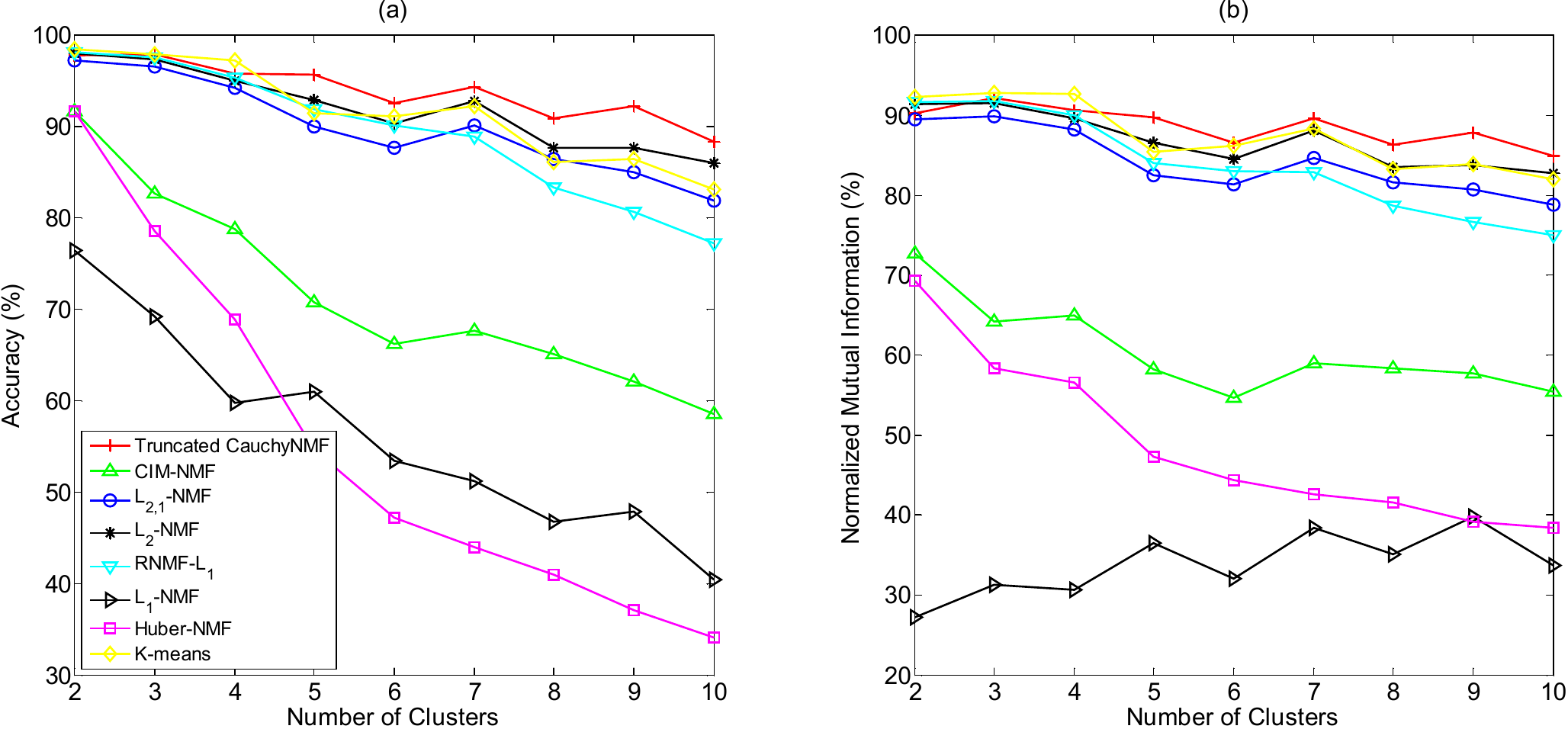}
\caption{The clustering performance, in terms of both accuracy and normalized mutual information, of Truncated CauchyNMF, CIM-NMF, Huber-NMF, $L_{2,1}$-NMF, RNMF-$L_1$, $L_2$-NMF, $L_1$-NMF, and K-means on the Caltech101 dataset with the number of clusters varying from $2$ to $10$.}
\label{fig5-11}
\end{figure}

Figure \ref{fig5-11} presents the accuracy and normalized mutual information versus cluster numbers of different NMF models. Truncated CauchyNMF significantly outperforms other approaches. As the number of categories increases, the accuracy achieved by other NMF models decreases quickly, while Truncated CauchyNMF maintains a strong subspace learning ability. We can see from the figure that Truncated CauchyNMF is more robust to the object variations compared to other models.

Note that, in all above experiments, we optimized the Truncated CauchyNMF and the other NMF models with different types of algorithms. However, the high performance is not due to the optimization algorithm. To study this point, we applied the Nesterov based HQ algorithm to optimize the representative NMF models and compared their clustering performance on the AR dataset. The results show that Truncated CauchyNMF consistently outperforms the other NMF models. See the supplementary materials for detailed discussions.

\section{Conclusion}\label{sec:chapter6}
This paper proposes a Truncated CauchyNMF framework for learning subspaces from corrupted data. We propose a Truncated Cauchy loss which can simultaneously and appropriately model both moderate and extreme outliers, and develop a novel Truncated CauchyNMF model. We theoretically analyze the robustness of Truncated CauchyNMF by comparing with a family of NMF models, and provide the performance guarantees of Truncated CauchyNMF. Considering that the objective function is neither convex nor quadratic, we optimize Truncated CauchyNMF by using half-quadratic programming and alternately updating both factor matrices. We experimentally verify the robustness and effectiveness of our methods on both synthetic and natural datasets and confirm that Truncated CauchyNMF are robust for learning subspace even when half the data points are contaminated.


%



\ifCLASSOPTIONcompsoc
  \section*{Acknowledgments}
\else
  \section*{Acknowledgment}
\fi
The work was supported in part by the Australian Research Council Projects FL-170100117, DP-180103424, DP-140102164, LP-150100671, and the National Natural Science Foundation of China under Grant 61502515.

\ifCLASSOPTIONcaptionsoff
  \newpage
\fi




\begin{thebibliography}{1}

\bibitem{bib2}
A.~Baccini, P.~Besse, and A.~Falguerolles, ``A $L_1$-norm PCA and a  Heuristic Approach,'' \emph{Ordinal and Symbolic Data Analysis}, pp. 359-368, Springer, 1996.
\bibitem{bib3}
P.~L.~Bartlett and S.~Mendelson, ``Rademacher and Gaussian Complexities: Risk Bounds and Structural Results,'' \emph{Journal of Machine Learning Research}, vol. 3, pp. 463-482, 2003.
\bibitem{bib4}
S.~P.~Boyd and L.~Vandenberghe, \emph{Convex Optimization}, Cambridge University Press, 2004.
\bibitem{bib5}
D.~Cai, X.~He, and J.~Han, ``Graph Regularized Non-negative Matrix Factorization for Data Representation,'' \emph{IEEE Transactions on Pattern Analysis and Machine Intelligence}, vol. 33, no. 8, pp. 1548-1560, 2011.
\bibitem{bib6}
A.~L.~Cauchy, ``On the Pressure or Tension in a Solid Body,'' \emph{Exercices de Math¨¦matiques}, vol. 2, no. 42, 1827.
\bibitem{bib8}
C.~Ding, T.~Li, and M.~I.~Jordan, ``Convex and Semi-non-negative Matrix Factorizations,'' \emph{IEEE Transactions on Pattern Analysis and Machine Intelligence}, vol. 32, no. 1, pp. 45-55, Jan. 2010.
\bibitem{bib9}
D.~Donoho, \emph{Breakdown Properties of Multivariate Location Estimators}, Qualifying paper, Harvard University, Cambridge MA, 1982.
\bibitem{bib10}
L.~Du, X.~Li, and Y.~D.~Shen, ``Robust Non-negative Matrix Factorization via Half-Quadratic Minimization,'' \emph{in Proceedings of IEEE 12th International Conference on Data Mining}, 2012, pp. 201-210.
\bibitem{bib13}
D.~Geman and C.~Yang, ``Nonlinear Image Recovery with Half-quadratic Regularization,'' \emph{IEEE Transactions on Image Processing}, vol. 4, no. 7, pp. 932-946, 1995.
\bibitem{bib16}
N.~Guan, D.~Tao, Z.~Luo, and B.~Yuan, ``NeNMF: An Optimal Gradient Method for Non-negative Matrix Factorization,'' \emph{IEEE Transactions on Signal Processing}, vol. 60, no. 6, pp. 2882-2898, 2012.
\bibitem{bib17}
N.~Guan, D.~Tao, Z.~Luo, and J.~Shawe-Taylor, ``MahNMF: Manhattan Non-negative Matrix Factorization,'' \emph{Journal of Machine Learning Research}, arXiv:1207.3438v1, 2012.
\bibitem{bib18}
A.~B.~Hamza and D.~J.~Brady, ``Reconstruction of Reflectance Spectra Using Robust Non-negative Matrix Factorization,'' \emph{IEEE Transactions on Signal Processing}, vol. 54, no. 9, pp. 3637-3642, 2006.
\bibitem{bib21}
H.~Hotelling, ``Analysis of a Complex of Statistical Variables into Principal Components,'' \emph{Journal of Educational Psychology}, vol. 24, pp. 417-441, 1933.
\bibitem{bib24}
D.~Kong, C.~Ding, and H.~Huang, ``Robust Non-negative Matrix Factorization using $L_{2,1}$-norm,'' \emph{in Proceedings of the 20th ACM International Conference on Information and Knowledge Management}, 2011, pp. 673-682.
\bibitem{bib25}
E.~Y.~Lam, ``Non-negative Matrix Factorization for Images with Laplacian Noise,'' \emph{in IEEE Asia Pacific Conference on Circuits and Systems}, 2008, pp. 798-801.
\bibitem{bib26}
D.~D.~Lee and H.~S.~Seung, ``Learning the Parts of Objects by Non-negative Matrix Factorization,'' \emph{Nature}, vol. 401, no. 6755, pp. 788-791, Oct. 1999.
\bibitem{bib27}
D.~D.~Lee and H.~S.~Seung, ``Algorithms for Non-negative Matrix Factorization,'' \emph{in Advances in Neural Information Process Systems}, 2001, pp. 556-562.
\bibitem{bib28}
C.~J.~Lin, ``Projected Gradient Methods for Non-negative Matrix Factorization,'' \emph{Neural Computation}, vol. 19, no. 10, pp. 2756-2779, Oct. 2007.
\bibitem{bib30}
W.~Liu, P.~P.~Pokharel, and J.~C.~Principe, ``Correntropy: Properties and Applications in Non-Gaussian Signal Processing,'' \emph{IEEE Transactions on Signal Processing}, vol. 55, no. 11, pp. 5286-5298, 2007.
\bibitem{bib31}
T.~Liu and D.~Tao, ``On the Robustness and Generalization of Cauchy Regression,'' \emph{IEEE International Conference on Information Science and Technology}, pp. 100-105, 26-28 April, 2014.
\bibitem{bib32}
J.~MacQueen, ``Some Methods for Classification and Analysis of Multivariate Observations,'' \emph{in Proceedings of the Fifth Berkeley Symposium on Mathematical Statistics and Probability}, vol. 1, no. 281-297, pp. 14, 1967.
\bibitem{bib33}
A.~Martinez and R.~Benavente, ``The AR Face Database,'' \emph{CVC Technical Report}, NO. 24, 1998.
\bibitem{bib35}
Y.~E.~Nesterov, ``A Method of Solving a Convex Programming Problem with Convergence Rate $O(1/k^2)$,'' \emph{Soviet Mathematics Doklady}, vol. 27, no. 2, pp. 372-376, 1983.
\bibitem{bib36}
M.~Nikolova and R.~H.~Chan, ``The Equivalence of Half-quadratic Minimization and the Gradient Linearization Iteration,'' \emph{IEEE Transactions on Image Processing}, vol. 16, no. 6, pp. 1623-1627, Jun. 2007.
\bibitem{bib38}
V.~P.~Pauca, F.~Shahnaz, M.~W.~Berry, and R.~J.~Plemmons, ``Text Mining using Non-Negative Matrix Factorization,'' \emph{in 4th SIAM International Conference on Data Mining}, 2004, pp. 452-456.
\bibitem{bib39}
V.~Pauca, J.~Piper, and R.~Plemmons, ``Non-negative Matrix Factorization for Spectral Data Analysis,'' \emph{Linear Algebra and its Applications}, vol. 416, no. 1, pp. 29-47, Jul. 2006.
\bibitem{bib42}
F.~Samaria and A.~Harter, ``Parameterisation of a Stochastic Model for Human Face Identification,'' \emph{in IEEE Workshop on Application and Computer Vision}, Sarasota, FL, 1994, pp. 138-142.
\bibitem{bib44}
R.~Sandler and M.~Lindenbaum, ``Non-negative Matrix Factorization with Earth Mover's Distance Metric for Image Analysis,'' \emph{IEEE Transactions on Pattern Analysis and Machine Intelligence}, vol. 33, no. 8, pp. 1590-1602, Jan. 2011.
\bibitem{bib48}
L.~Zhang, Z.~Chen, M.~Zheng, and X.~He, ``Robust Non-negative Matrix Factorization,'' \emph{Frontiers of Electrical and Electronic Engineering in China}, vol. 6, no. 2, pp. 192-200, Feb. 2011.
\bibitem{bib49}
T.~Zhang, ``Covering Number Bounds of Certain Regularized Linear Function Classes,'' \emph{Journal of Machine Learning Research}, vol. 2, pp. 527-550, 2002.
\bibitem{bib50}
Y.~Jia and T.~Darrell, ``Heavy-tailed Distances for Gradient Based Image Descriptors,'' \emph{in Advances in Neural Information Systems}, 2011.
\bibitem{bib51}
F.~Nagy, ``Parameter Estimation of the Cauchy Distribution in Information Theory Approach,'' \emph{Journal of Universal Computer Science}, vol. 12, no. 9, pp. 1332-1344, 2006.
\bibitem{bib52}
L.~K.~Chan, ``Linear Estimation of the Location and Scale Parameters of the Cauchy Distribution Based on Sample Quantiles,'' \emph{Journal of the American Statistical Association}, vol. 65, no. 330, 1970.
\bibitem{bib53}
G.~Cane, ``Linear Estimation of Parameters of the Cauchy Distribution Based on Sample Quantiles,'' \emph{Journal of the American Statistical Association}, vol. 69, no. 345, pp. 243-245, 1974.
\bibitem{bib56}
P.~Chen, N.~Wang, N.~L.~Zhang, and D.~Y.~Yeung, ``Bayesian Adaptive Matrix Factorization with Automatic Model Selection,'' \emph{in Proceedings of the IEEE Computer Society Conference on Computer Vision and Pattern Recognition}, Boston, MA, pp. 7-12, June 2015.
\bibitem{bib58}
Y.~Nesterov, ``Smooth minimization of non-smooth functions,'' \emph{Mathematical Programming}, vol. 103, no. 1, pp. 127-152, Dec. 2005.
\bibitem{bib59}
G.~H.~Golub and C.~Reinsch, ``Singular value decomposition and least squares solutions,'' \emph{Numerische mathematik}, vol. 14, no. 5, pp. 403-420, 1970.
\bibitem{bib60}
J.~P.~Brunet, P.~Tamayo, T.~R.~Golub, and J.~P.~Mesirov,``Metagenes and molecular pattern discovery using matrix factorization,'' \emph{Proceedings of the national academy of sciences}, vol. 101, no. 12, pp. 4164-4169, 2004.
\bibitem{bib61}
Y.~Jia, E.~Shelhamer, J.~Donahue, S.~Karayev, J.~Long, R.~Girshick, S.~Guadarrama, and T.~Darrell,``Caffe: Convolutional architecture for fast feature embedding,'' \emph{In Proceedings of the ACM International Conference on Multimedia}, pp. 675-678, 2014.
\bibitem{bib62}
N.~K.~Logothetis and D.~L.~Sheinberg, ``Visual Object Recognition'', \emph{Annual Review of Neuroscience}, vol. 19, pp. 577-621, 1996.
\bibitem{bib64}
E.~Wachsmuth, M.~W.~Oram, and D.~I.~Perrett, ``Recognition of Objects and Their Component Parts: Responses of Single Units in the Temporal Cortex of the Macaque'', \emph{Cerebral Cortex}, vol. 4, pp. 509-522, 1994.
\bibitem{bib65}
A.~Krizhevsky, I.~Sutskever, and G.~E.~Hinton, ``Imagenet classification with deep convolutional neural networks,'' \emph{In Advances in neural information processing systems}, pp. 1097-1105, 2012.
\bibitem{bib66}
P.~J.~Huber, \emph{Robust statistics}, Springer Berlin Heidelberg, 2011.
\bibitem{bib67}
L.~Fei-Fei, R.~Fergus, and P.~Perona, ``Learning Generative Visual Models From Few Training Examples: An Incremental Bayesian Approach Tested on $101$ Object Categories,'' \emph{Computer Vision and Image Understanding}, vol. 106, no. 1, pp. 59-70, 2007.
\bibitem{bib69}
J.~Wright, A.~Y.~Yang, A.~Ganesh A, S.~S.~Sastry, Y.~Ma, ``Robust Face Recognition via Sparse Representation,'' \emph{IEEE Transactions on Pattern Analysis and Machine Intelligence}, vol. 31, no. 2, pp. 210-227, 2009.
\bibitem{bib68}
Z.~Zhou, A.~Wagner, H.~Mobahi, J.~Wright, and Y.~Ma, ``Face Recognition with Contiguous Occlusion Using Markov Random Fields,'' \emph{In Proceedings of International Conference on Computer Vision}, pp. 1050-1057, 2009.
\bibitem{bib70}
H.~Gao, F.~Nie, W.~Cai, and H.~Huang, ``Robust Capped Norm Nonnegative Matrix Factorization,'' \emph{In Proceedings of the 24th ACM International Conference on Information and Knowledge Management}, Oct. 19-23, Melbourne, Austrialia, 2015.
\bibitem{bib71}
C.~Bhattacharyya, N.~Goyal, R.~Kannan, and J.~Pani, ``Non-negative Matrix Factorization under Heavy Noise,'' \emph{In Proceedings of the 33th International Conference on Machine Learning}, New York, NY, USA, 2016.
\bibitem{bib72}
Q.~Pan, D.~Kong, C.~Ding, and B.~Luo, ``Robust Non-Negative Dictionary Learning,'' \emph{In Proceedings of the Twenty-Eighth AAAI Conference on Artificial Intelligence}, pp. 2027-2033, 2014.
\bibitem{bib73}
N.~Gillis and R.~Luce, ``Robust Near-Separable Nonnegative Matrix Factorization Using Linear Optimization,'' \emph{Journal of Machine Learning Research}, vol. 15, pp. 1249-1280, 2014.
\end{thebibliography}
%

%
\begin{IEEEbiography}[{\includegraphics[width=1in,height=1.25in,clip,keepaspectratio]{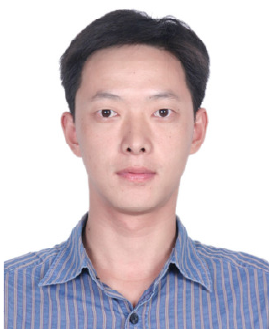}}]{Naiyang Guan} is currently an Associate Professor with the College of Computer, at The National University of Defense Technology of China. He received the BS, MS, and PhD degree from The National University of Defense Technology. His research interests include machine learning, computer vision, and data mining. He has authored and co-authored 20+ research papers including IEEE T-NNLS, T-IP, T-SP, Neurocomputing, Genes, BMC Genomics, ICDM, IJCAI, ECCV, PAKDD, ICONIP, ICASSP, IJCNN.
\end{IEEEbiography}

\begin{IEEEbiography}[{\includegraphics[width=1in,height=1.25in,clip,keepaspectratio]{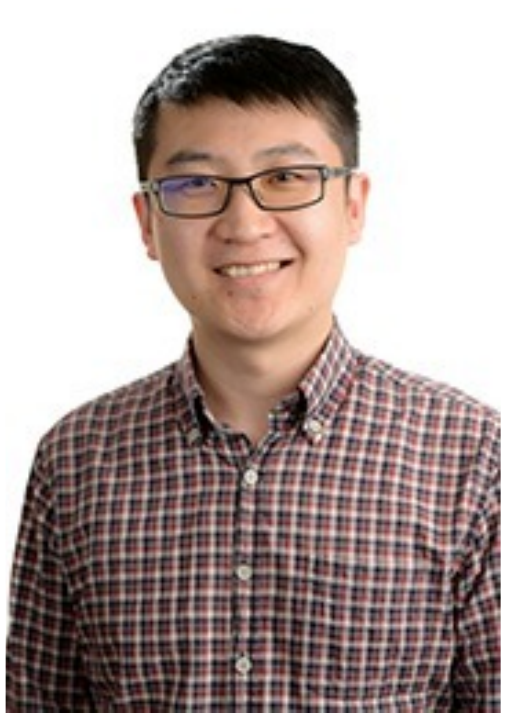}}]{Tongliang Liu} is currently a Lecturer with the School of Information Technologies and the Faculty of Engineering and Information Technologies, and a core member in the UBTECH Sydney AI Centre, at The University of Sydney. He received the BEng degree in electronic engineering and information science from the University of Science and Technology of China, and the PhD degree from the University of Technology Sydney. His research interests include statistical learning theory, computer vision, and optimisation. He has authored and co-authored 30+ research papers including IEEE T-PAMI, T-NNLS, T-IP, ICML, CVPR, and KDD.
\end{IEEEbiography}

\begin{IEEEbiography}[{\includegraphics[width=1in,height=1.25in,clip,keepaspectratio]{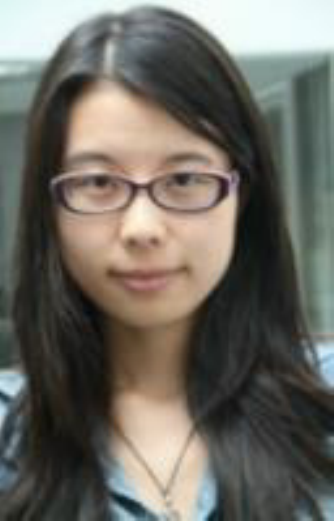}}]{Yangmuzi Zhang}
received her B.S. in electronics and information engineering from Huazhong University of Science and Technology in 2011. She obtained her Ph.D. in computer vision in the University of Maryland at College Park in 2016. She is currently working at Google.  Her research interests are computer vision and machine learning.
\end{IEEEbiography}

\begin{IEEEbiography}[{\includegraphics[width=1in,height=1.25in,clip,keepaspectratio]{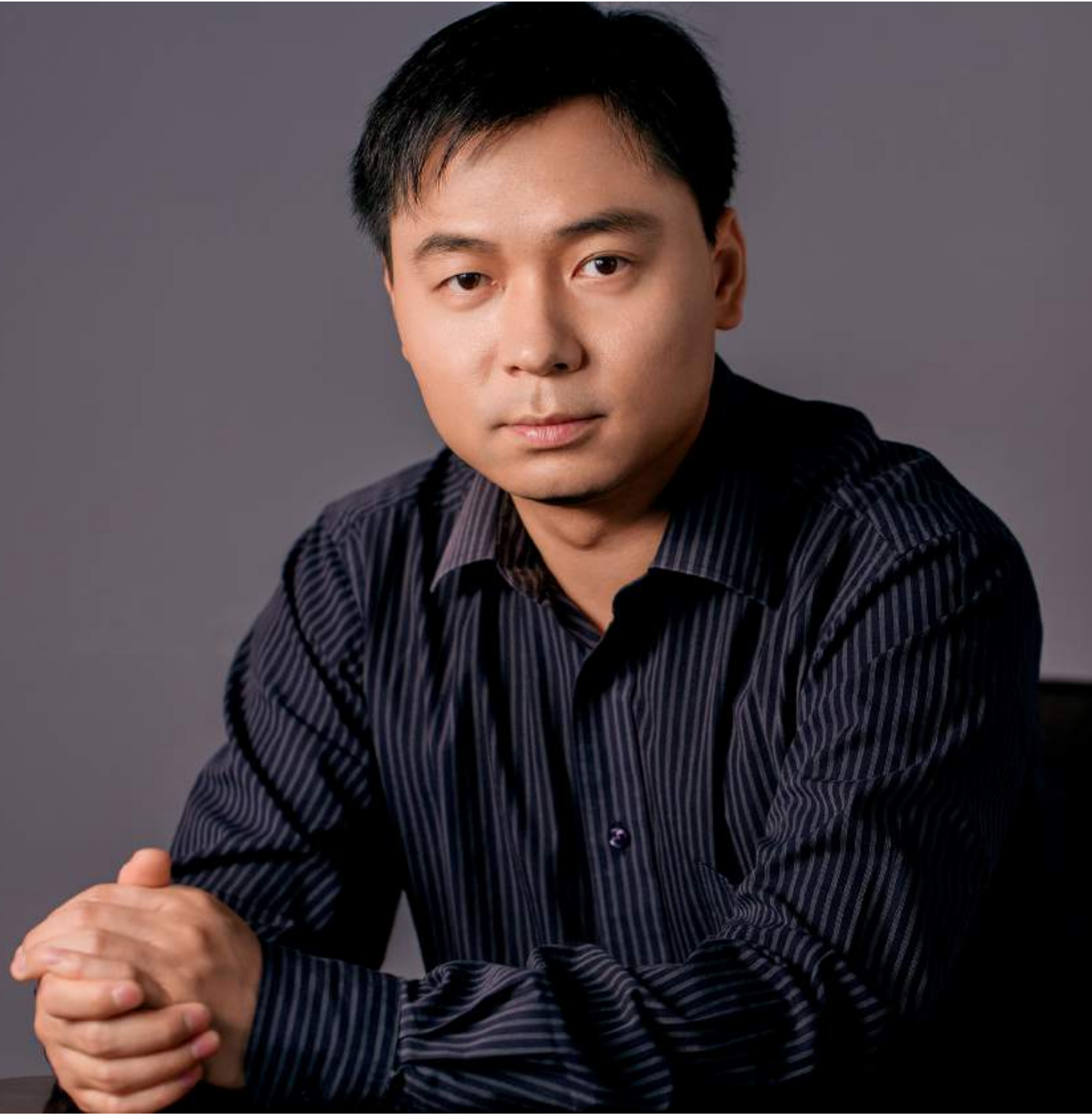}}]{Dacheng Tao}
(F'15) is Professor of Computer Science and ARC Laureate Fellow in the School of Information Technologies and the Faculty of Engineering and Information Technologies, and the Inaugural Director of the UBTECH Sydney Artificial Intelligence Centre, at the University of Sydney. He mainly applies statistics and mathematics to Artificial Intelligence and Data Science. His research interests spread across computer vision, data science, image processing, machine learning, and video surveillance. His research results have expounded in one monograph and 500+ publications at prestigious journals and prominent conferences, such as IEEE T-PAMI, T-NNLS, T-IP, JMLR, IJCV, NIPS, ICML, CVPR, ICCV, ECCV, ICDM; and ACM SIGKDD, with several best paper awards, such as the best theory/algorithm paper runner up award in IEEE ICDM'07, the best student paper award in IEEE ICDM'13, the distinguished student paper award in the 2017 IJCAI, the 2014 ICDM 10-year highest-impact paper award, and the 2017 IEEE Signal Processing Society Best Paper Award. He received the 2015 Australian Scopus-Eureka Prize, the 2015 ACS Gold Disruptor Award and the 2015 UTS Vice-Chancellor's Medal for Exceptional Research. He is a Fellow of the IEEE, OSA, IAPR and SPIE.
\end{IEEEbiography}

\begin{IEEEbiography}[{\includegraphics[width=1in,height=1.25in,clip,keepaspectratio]{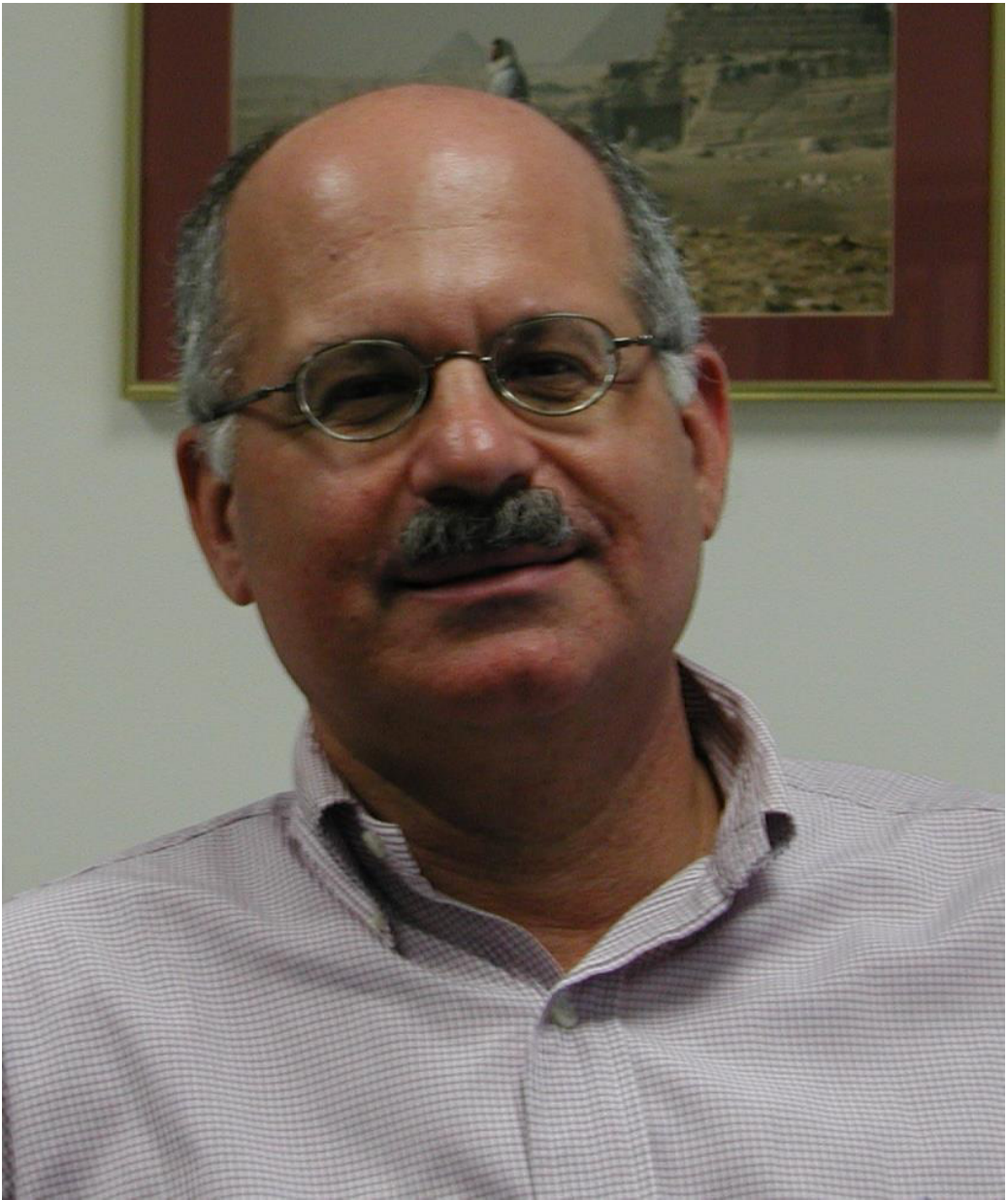}}]{Larry S. Davis}
received his B.A. from Colgate University in 1970 and his M. S. and Ph. D. in Computer Science from the University of Maryland in 1974 and 1976 respectively. From 1977-1981 he was an Assistant Professor in the Department of Computer Science at the University of Texas, Austin. He returned to the University of Maryland as an Associate Professor in 1981. From 1985-1994 he was the Director of the University of Maryland Institute for Advanced Computer Studies. He was Chair of the Department of Computer Science from 1999-2012. He is currently a Professor in the Institute and the Computer Science Department, as well as Director of the Center for Automation Research. He was named a Fellow of the IEEE in 1997 and of the ACM in 2013.
\end{IEEEbiography}







\end{document}